\RequirePackage{fix-cm}
\documentclass[twocolumn]{svjour3}          
\smartqed  
\usepackage{framed,multirow}
\usepackage[justification=centering]{caption}
\usepackage{graphicx}
\usepackage{subfigure}
\usepackage{caption}
\usepackage{epstopdf}
\usepackage{amsfonts}
\usepackage{amsmath}
\usepackage{supertabular}
\usepackage{array}
\usepackage{color}
\usepackage{comment}
\usepackage{url}
\usepackage{breakurl}
\newcommand{\tabincell}[2]{\begin{tabular}{@{}#1@{}}#2\end{tabular}}
\newcommand{\selftopcaption}{%
\setlength{\abovecaptionskip}{0pt}%
\setlength{\belowcaptionskip}{10pt}%
\caption}
\begin{document}

\title{A Survey of Deep Facial Attribute Analysis
}


\author{Xin Zheng         \and
        Yanqing Guo*       \and
        Huaibo Huang      \and
        Yi Li             \and
        Ran He            \and
}


\institute{Xin Zheng \at
              School of Information and Communication Engineering, Dalian University of Technology, Dalian 116024, China\\
              \email{zhengxin@mail.dlut.edu.cn}           
           \and
           Yanqing Guo* \at
              School of Information and Communication Engineering, Dalian University of Technology, Dalian 116024, China\\
              \email{guoyq@dlut.edu.cn}
           \and
           Huaibo Huang \at
              National Laboratory of Pattern Recognition, CASIA\\
              Center for Research on Intelligent Perception and Computing, CASIA\\
              University of Chinese Academy of Sciences, Beijing 100190, China
              \email{huaibo.huang@cripac.ia.ac.cn}
           \and
           Yi Li \at
              National Laboratory of Pattern Recognition, CASIA\\
              Center for Research on Intelligent Perception and Computing, CASIA\\
              University of Chinese Academy of Sciences, Beijing 100190, China
              \email{yi.li@cripac.ia.ac.cn}
           \and
           Ran He \at
              National Laboratory of Pattern Recognition, CASIA\\
              Center for Research on Intelligent Perception and Computing, CASIA\\
              Center for Excellence in Brain Science and Intelligence Technology, CAS\\
              University of Chinese Academy of Sciences, Beijing 100190, China
              \email{rhe@nlpr.ia.ac.cn}
}
\date{Received: date / Accepted: date}

\maketitle

\begin{abstract}
Facial attribute analysis has received considerable attention when deep learning techniques made remarkable breakthroughs in this field over the past few years. Deep learning based facial attribute analysis consists of two basic sub-issues: facial attribute estimation (FAE), which recognizes whether facial attributes are present in given images, and facial attribute manipulation (FAM), which synthesizes or removes desired facial attributes. In this paper, we provide a comprehensive survey of deep facial attribute analysis from the perspectives of both estimation and manipulation. First, we summarize a general pipeline that deep facial attribute analysis follows, which comprises two stages: data preprocessing and model construction. Additionally, we introduce the underlying theories of this two-stage pipeline for both FAE and FAM. Second, the datasets and performance metrics commonly used in facial attribute analysis are presented. Third, we create a taxonomy of state-of-the-art methods and review deep FAE and FAM algorithms in detail. Furthermore, several additional facial attribute related issues are introduced, as well as relevant real-world applications. Finally, we discuss possible challenges and promising future research directions.
\end{abstract}

\section{Introduction}
\label{intro}
Facial attributes represent intuitive semantic features that describe human-understandable visual properties of face images, such as \emph{smiling}, \emph{eyeglasses}, and \emph{mustache}. Therefore, as vital information of faces, facial attributes have contributed to numerous real-world applications, e.g., face verification \cite{kumar2009attribute,berg2013poof,song2014verification,zhang2018demeshnet,chen2018unconstrained}, face recognition \cite{he2018wasserstein,shi2015person_reid,he2018learning,song2018adversarial,rao2018learning}, face retrieval \cite{li2015twobrtdonestone,nguyen2018large,fang2018attribute,toderici2010ethnicity}, and face image synthesis \cite{huang2018introvae,cao2018load,huang2018variational,song2018geometry,egger2018occlusion}. Facial attribute analysis, aiming to build a bridge between human-understandable visual descriptions and abstract feature representations required by real-world computer vision tasks, has attracted increasing attention and has become a hot research topic. Recently, the development of deep learning techniques has made excellent progress in learning abstract feature representations, leading to significant performance improvements of the current algorithms in the field of deep facial attribute analysis.

\begin{figure}[t]
\centering
\subfigure[FAE]
{\includegraphics[width=0.45\linewidth]{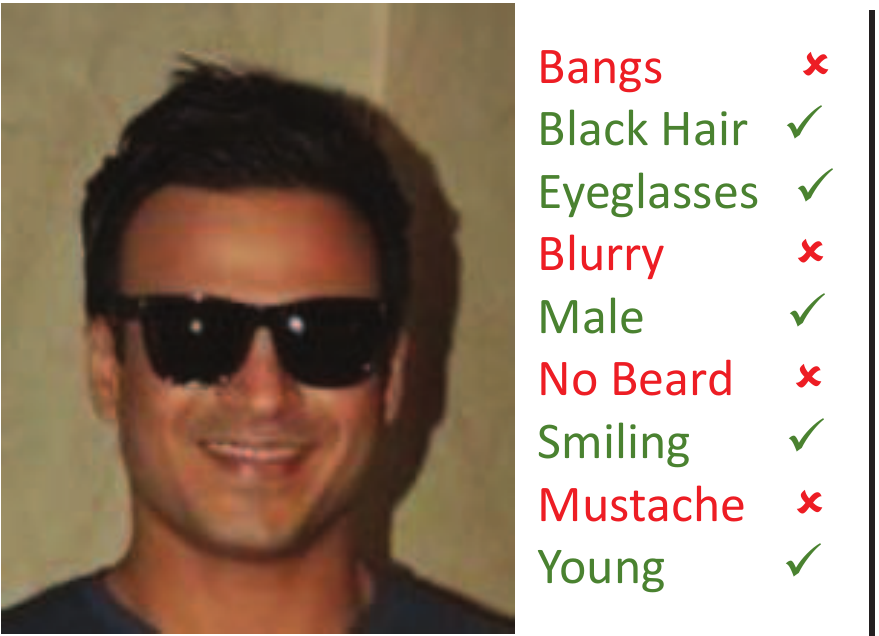}}
\subfigure[FAM]
{\includegraphics[width=0.45\linewidth]{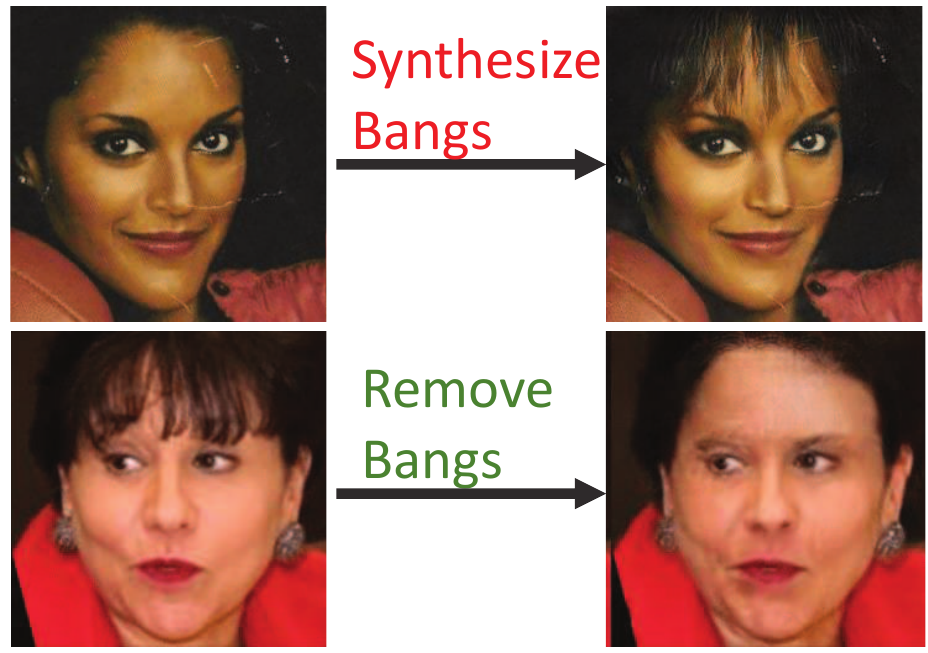}}
\caption{Illustrations of the two sub-issues in deep facial attribute analysis, i.e., FAE and FAM
((a) comes from CelebA dataset \cite{liu2015deep}, and (b) comes from \cite{xiao2018elegant}).}
\label{FAAGeneral}
\end{figure}

Deep facial attribute analysis mainly consists of two sub-issues: facial attribute estimation (FAE) and facial attribute manipulation (FAM). Given a face image, FAE trains attribute classifiers to recognize whether a specific facial attribute is present, and FAM modifies face images to synthesize or remove desired attributes by constructing generative models. We provide concise illustrations of these two sub-issues in Fig.~\ref{FAAGeneral}.

Deep FAE methods can generally be categorized into two groups: part-based methods and holistic methods. Part-based FAE methods first locate the positions of facial attributes and then extract features according to the obtained localization cues for the subsequent attribute prediction. According to the different schemes for locating facial attributes, part-based methods can be further classified into two subcategories: separate auxiliary localization based methods and end-to-end localization based methods. Specifically, separate auxiliary localization based FAE methods seek help from existing part detectors or auxiliary localization algorithms, e.g., facial key point detection \cite{mahbub2018segment,wu2017facial} and semantic segmentation \cite{kalayeh2017improving,gonzalez2018semantic}. Then, corresponding features from different positions can be extracted for further estimation. Note that the localization and the estimation are performed in a separate and independent manner. On the contrary, end-to-end localization based methods exploit the locations of attributes and predict their presence simultaneously in end-to-end frameworks. In contrast to part-based methods, holistic methods focus more on learning attribute relationships and estimating facial attributes in a unified framework without any additional localization modules. By assigning shared and specific attribute learning to different layers of networks, holistic methods model correlations and distinctions among facial attributes to explore the complementary information. During this process, holistic FAE algorithms resort to additional prior or auxiliary information, such as attribute grouping or identity information \cite{cao2018partially}, to customize their network architectures.

\begin{figure*}[t]
\centering
\includegraphics[width=0.8\linewidth]{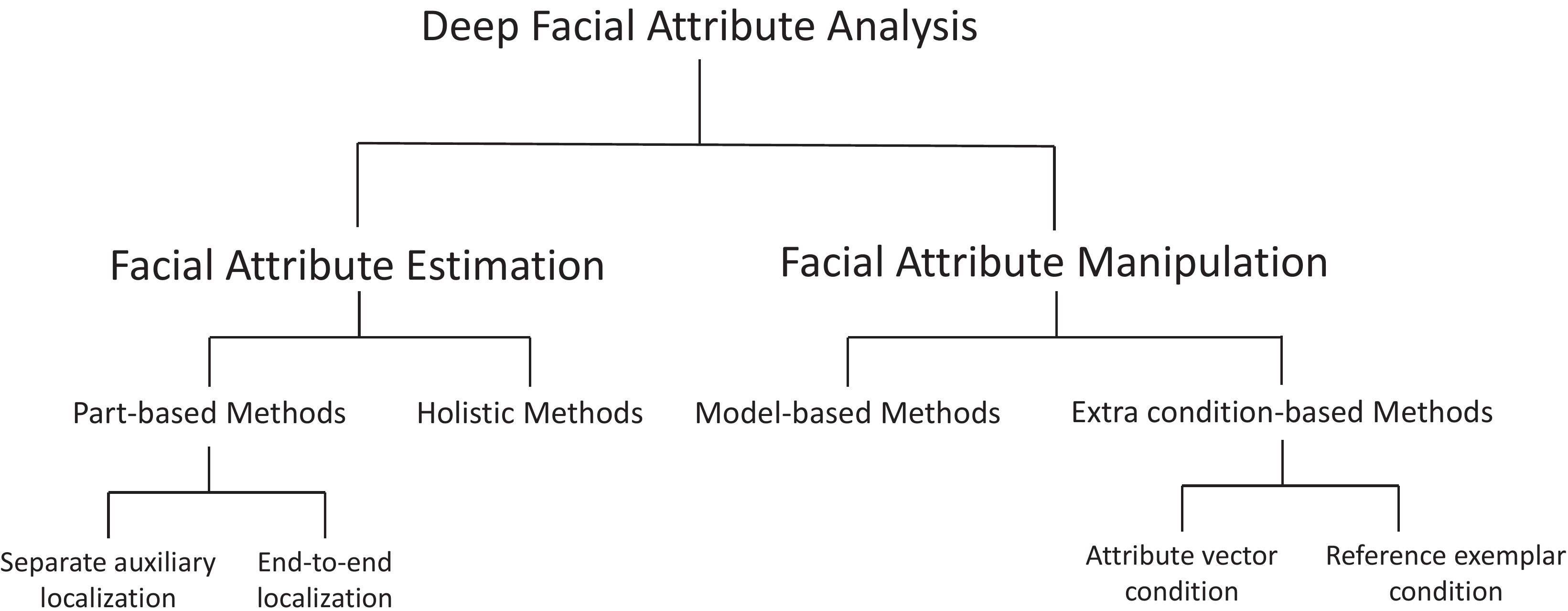}
\caption{Tree diagram for diverse categories of deep facial attribute analysis algorithms.}
\label{Tree diagram of FAA}
\end{figure*}
Deep FAM methods are mainly constructed based on generative models, of which generative adversarial networks (GANs) \cite{goodfellow2014gan,mirza2014conditional,chen2016infogan} and variational autoencoders (VAEs) \cite{kingma2013auto_bayes,huang2018introvae,huang2018variational} serve as the backbones. Furthermore, deep FAM algorithms can be divided into two groups: model-based methods and extra condition-based methods, where the main difference between them is whether extra conditions are introduced. Model-based methods construct a model without any extra conditional inputs and learn a set of model parameters that only correspond to one attribute during a single training process. Thus, when editing another attribute, another training process needs to be executed in the same way. In this case, multiple attribute manipulations correspond to multiple training processes, resulting in expensive computation costs. In contrast, extra condition-based methods take extra attribute vectors or reference images as input conditions, and they can alter multiple attributes simultaneously by changing the corresponding values of attribute vectors or taking multiple exemplars with distinct attributes as references. Specifically, given an original image, an extra conditional attribute vector, such as a one-hot vector indicating the presence of the attribute, is concatenated with the latent original image codes. By comparison, extra conditional reference exemplars exchange specific attributes with the original image in the framework of image-to-image translation. Note that these reference images do not need to have the same identity as the original image. Hence, rather than merely altering the values of attribute vectors to edit facial attributes, attribute transfer based on reference images can discover more specific details of references and yield more faithful facial attribute images \cite{zhou2017genegan,xiao2018elegant,ma2018exemplar}. Due to more abundant facial details and more photorealistic performance of generated images, this type of method has attracted much attention of current researchers.
\begin{figure*}[t]
\centering
\includegraphics[width=0.95\linewidth]{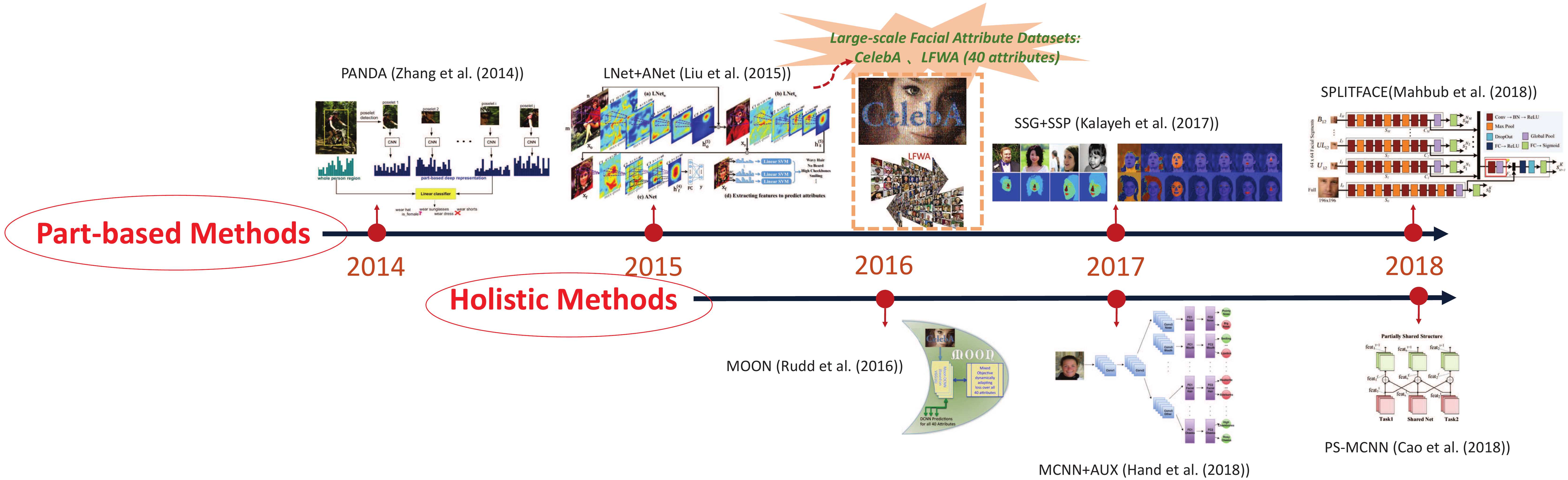}
\caption{The evolution of deep FAE methods (Best viewed by zooming in the electronic version).}
\label{Timeline_FAE}
\end{figure*}

\begin{figure*}[t]
\centering
\includegraphics[width=0.95\linewidth]{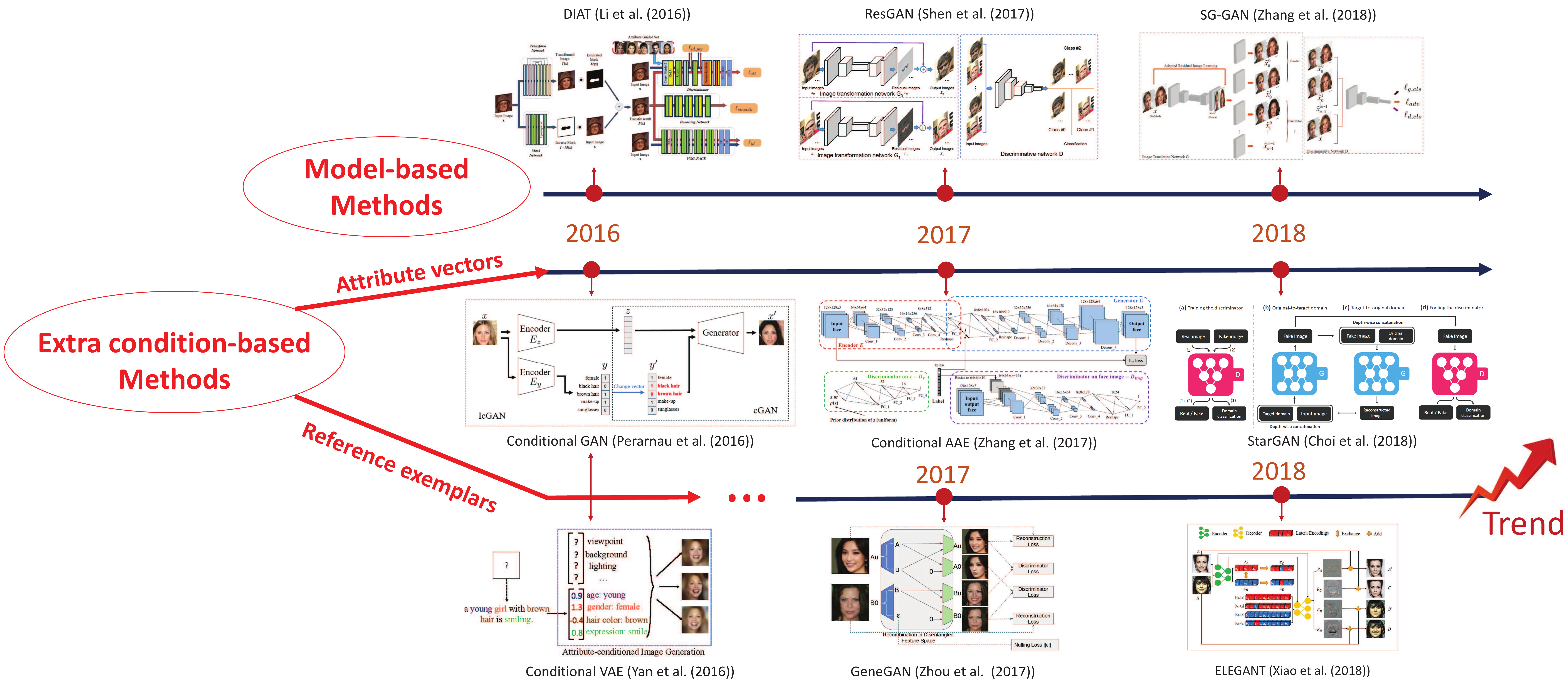}
\caption{The evolution of deep FAM methods (Best viewed by zooming in the electronic version).}
\label{Timeline_FAM}
\end{figure*}

In summary, we create a taxonomy of contemporary deep facial attribute analysis algorithms in a tree diagram in Fig.~\ref{Tree diagram of FAA}. Furthermore, aiming to summarize the progress in deep facial attribute analysis, milestones of both deep FAE and FAM methods are listed in Fig.~\ref{Timeline_FAE} and Fig.~\ref{Timeline_FAM}, respectively.

As shown in Fig.~\ref{Timeline_FAE}, part-based FAE methods and holistic FAE methods share two parallel routes. The study of deep FAE can be traced back to the earliest part-based work of Zhang et al. \cite{zhang2014panda}, who take the whole person images as inputs.
Then, LNet+ANet \cite{liu2015deep} pushes deep FAE into an independent research branch, where only face images are taken as inputs for merely estimating face-related attributes. In addition, two large-scale face datasets, i.e., CelebA and LFWA, with 40 labeled attributes, are released to provide data support for deep FAE methods. Then, part-based and holistic methods share joint development and success but have distinct directions and trends. Part-based methods extremely emphasize facial details for discovering localization cues \cite{kalayeh2017improving,mahbub2018segment}, whereas holistic methods incline to employ attribute relationships to customize networks for learning more discriminative features \cite{rudd2016moon,hand2017attributes,cao2018partially}.

We outline the development of deep FAM methods in Fig.~\ref{Timeline_FAM}. Note that, model-based methods and two types of extra condition-based methods have their own evolutionary processes, but all follow the advances in GANs or VAEs. The earliest deep FAM work DIAT \cite{li2016IDaware}, a model-based method, first attempts to utilize simple GANs to generate facial attributes. Meanwhile, conditional GAN \cite{perarnau2016invertible} and VAE \cite{yan2016attribute2image} dominate extra condition-based FAM methods by taking attribute vectors as conditions. Though extra attribute vector based methods have the remarkable advantage of changing multiple attributes simultaneously, they cannot guarantee that the remaining details that are irrelevant to manipulated attributes keep unchanged. Model-based methods can overcome this problem, but they cannot manipulate multiple attributes in a single training process. In light of these issues, methods conditioned on reference exemplars come into researchers' attention. They can balance the change of multiple interested attributes and the preservation of other irrelevant attributes; meanwhile, generate more photorealistic facial attribute images. Hence, exemplar-guided FAM methods are becoming a popular research trend.
\begin{figure}[t]
\centering
\includegraphics[width=1\linewidth]{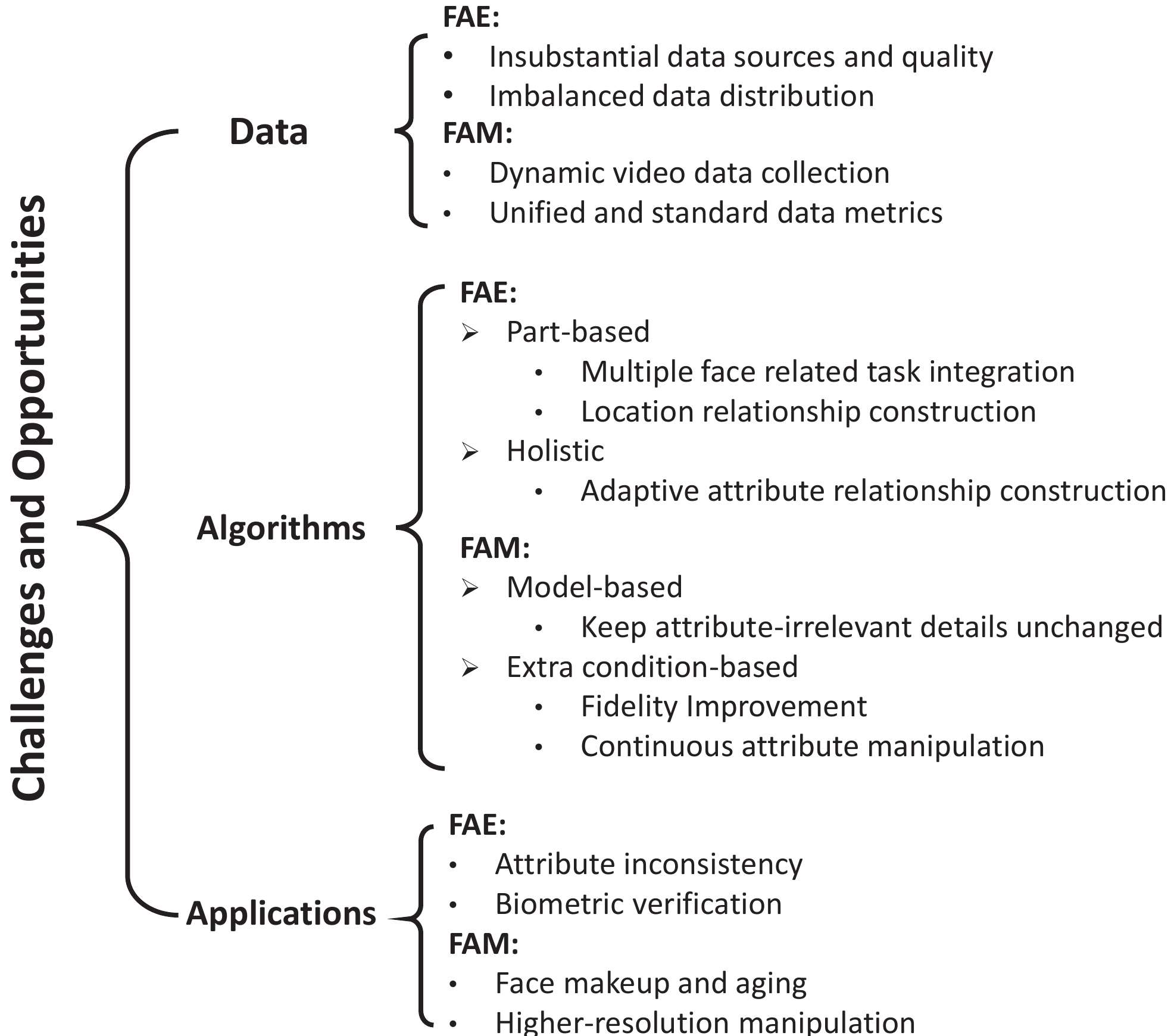}
\caption{Summary of challenges and opportunities in deep facial attribute analysis.}
\label{Challenges}
\end{figure}
Although a large number of methods achieve appealing performance in deep FAE and FAM methods, there are still several severe challenges for future deep facial attribute analysis. Therefore, we summarize these urgent challenges and analyze possible opportunities in terms of data, algorithms, and applications. The corresponding overview is described in Fig.~\ref{Challenges}.

First, from the perspective of data, contemporary deep FAE methods suffer from the problem of insufficient training data. The most commonly used two datasets come from celebrities or news \cite{liu2015deep}, where attribute types, illumination, views, and poses, all have significant differences from real-world data \cite{hand2018doing}. Therefore, future deep FAE models would have high demands for diverse data sources and excellent data quality (e.g., video data \cite{wang2016walk,hand2018predicting}). Future facial attribute images need to cover more real-world scenarios and wider-range attribute types. In this way, models can better capture representative features that conform to real-world data distribution. In addition, imbalanced data distribution of facial attribute images highlights in two aspects: the attribute category imbalance over a single dataset and the domain gaps between different training and testing datasets. The former called class-imbalance issue makes FAE models bias towards the majority samples and ignore the minority ones, resulting in the degraded performance in minority sample recognition. In contrast, the latter called domain adaption issue, which has not been fully explored in current deep FAE algorithms yet, is related to the generalization of models, especially when testing over unseen data.


Regarding the data challenges and opportunities in deep FAM, the rapid development of multimedia in the era of big data has given rise to rich video data. However, tracking and annotating facial attributes in videos is difficult because of spatial and temporal dynamics \cite{saito2017temporal}. Hence, video attribute manipulation is still a task to be addressed due to the lack of available training data. In addition, a large proportion of current algorithms evaluate the quality of their generated facial attribute images based on the visual fidelity \cite{li2016IDaware,perarnau2016invertible,zhang2017st,xiao2018elegant}. Because of the lack of established protocols and standards, such measurements might have adverse effects on the performance evaluation of deep FAM methods. Therefore, it would be a thorny problem to seek unified and standard data metric schemes that achieve both qualitative and quantitative analyses.

Second, from the perspective of algorithms, deep part-based FAE methods mainly focus on two aspects. The first is to integrate multiple face-related tasks, such as attribute estimation and face recognition, into a unified framework. In this way, the complementary information among different tasks could be fully exploited to improve all of them. For the second aspect, future part-based FAE algorithms are expected to discover more relationships among different attribute locations to handle in-the-wild data with complex environmental variations. For deep holistic FAE algorithms, current algorithms discover attribute relationships with the help of the prior information, e.g., human-made facial attribute groups. Such artificial partitions would limit the generalization ability of models. Hence, the critical challenge that holistic FAE methods face is to design adaptive attribute partition schemes for automatically exploring attribute relationships during the training processes.

With regard to the algorithm challenges and opportunities in deep FAM, model-based methods have a severe drawback: they cannot keep other attribute-irrelevant details unchanged as supervised information only comes from the target images with desired attributes. In terms of extra condition-based FAM methods, on the one hand, attribute vector based algorithms need to work harder to manipulate attributes continuously, where interpolation schemes might be a solution worth considering. On the other hand, future reference exemplar-based algorithms are expected to generate more diverse attribute styles in more faithful and photorealistic face images.

Finally, from the perspective of applications in deep FAE, face images with different viewpoints might have different attributes for the same person. It is possible that an attribute shown on the front face is not emphasized on the profile. This is called attribute inconsistency issue. By filtering more confident images to make the prediction \cite{lu2018attribute}, existing methods might neglect rich information in multi-view face images. Therefore, how to keep attributes from the same identity consistent, while taking full advantage of information for capturing features with multiple views are important questions for the future. Second, biometric verification \cite{hadid2007face,gunther20132013,fathy2015face,samangouei2017facial,trokielewicz2019iris} is a developing application for digital mobile devices to resist various attacks in the real world. Compared with full-face based biometric verification \cite{fathy2015face,gunther20132013}, facial attributes contain more detailed characteristics and can better facilitate active authentication. The main obstacles lie in the following two aspects: the first is to introduce facial attributes into the task of active authentication appropriately and efficiently \cite{samangouei2017facial}, and the second is to explore the available deep features and classifiers with the trade-off between the verification accuracy and mobile performance.

Regarding the application challenges and opportunities in deep FAM, facial makeup \cite{li2018beautygan,chang2018pairedcyclegan,cao2019makeup} and aging \cite{suo2010compositional,nhan2019automatic,liu2019attribute} have become hot topics in computer vision. The two tasks focus more on subtle facial details related to makeup and age attributes. Due to promising performance in mobile devices entertainment and identity-relevant verification, they have turned into crucial study branches independent of general deep FAM methods, and have shown significant potentials to facilitate more practical applications \cite{hu2018pose,lu2018conditional,songlx2018geometry}. In addition, contemporary deep FAM research only works well with a limited range of resolutions and under laboratory conditions. On the one hand, such a limitation leads to more difficult high-resolution and low-quality face image manipulation in real-world applications; on the other hand, it provides an opportunity to integrate face super-resolution into attribute manipulation \cite{lu2018attribute,dorta2018gan} in future research.

In addition, the relationships between deep FAE and FAM might contribute to improving both tasks. On the one hand, FAM is a vital scheme of data augmentation for FAE, where generated facial attribute images can significantly increase the amount of data to further relieve the overfitting issue. On the other hand, FAE can be a significant quantitative performance evaluation criterion for FAM, where the accuracy gap between real images and generated images can reflect the performance of deep FAM algorithms.

In this paper, we conduct an in-depth survey of facial attribute analysis based on deep learning, including FAE and FAM. The primary goal is to provide an overview of the two issues, and to highlight their respective strengths and weaknesses to provide newcomers prime skills. The remainder of this paper is organized as follows. In Section~\ref{section2}, we summarize a general two-stage pipeline that deep facial attribute analysis follows, including data preprocessing and model construction. The corresponding preliminary theories are also introduced for both FAE and FAM. In Section~\ref{section5}, we list commonly used publicly available facial attribute datasets and metrics. Section~\ref{section3} and Section~\ref{section4} provide detailed overviews of state-of-the-art deep FAE and FAM methods, as well as their advantages and disadvantages, respectively. Additional related issues, as well as challenges and opportunities, are discussed in Section~\ref{section6} and Section~\ref{section7}, respectively. Finally, we conclude this paper in Section~\ref{section8}.
\begin{figure*}[t]
\centering
\includegraphics[width=0.9\linewidth]{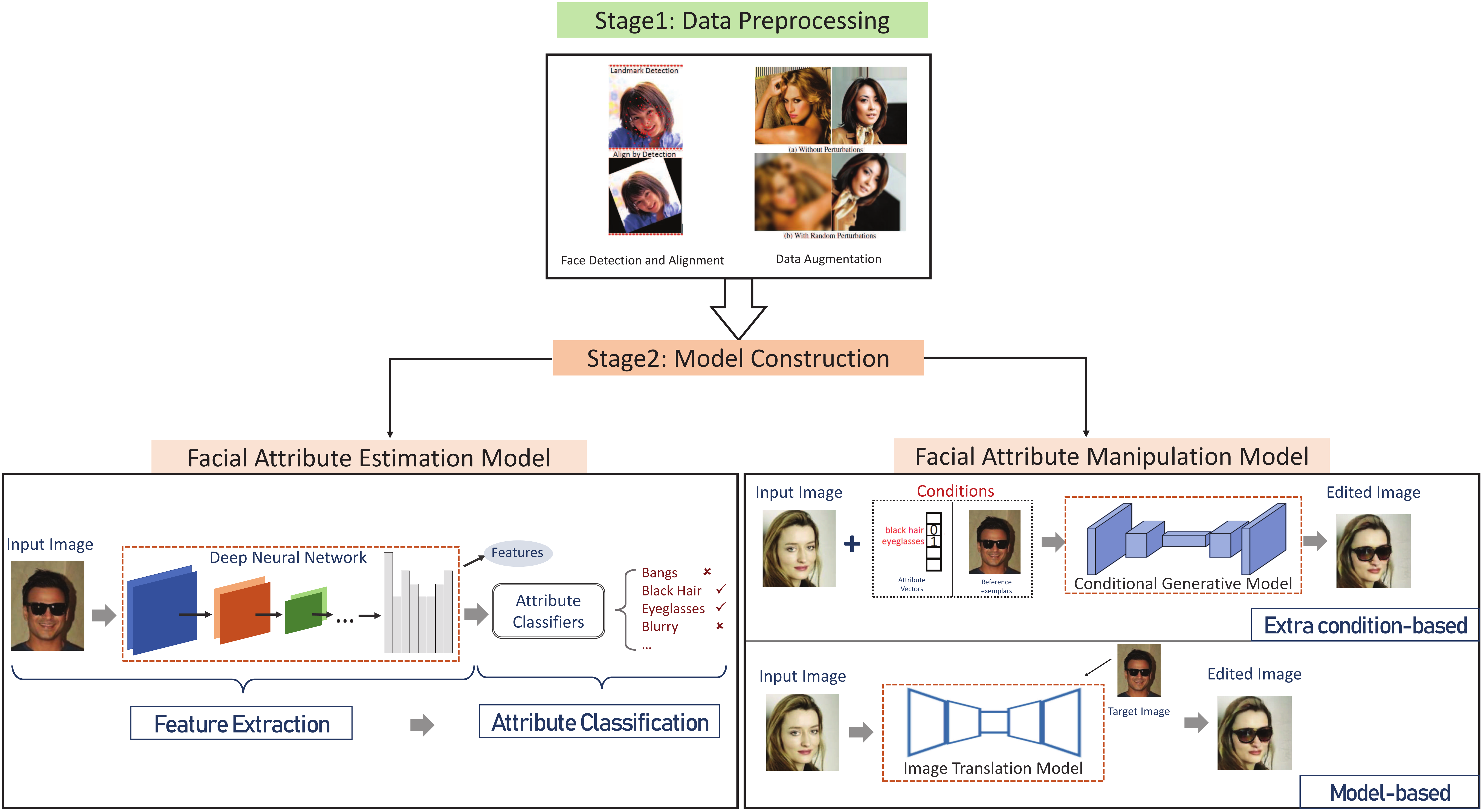}
\caption{Two-stage pipeline of deep facial attribute analysis (Face images above come from \cite{li2018landmark,GuntherRB17Affact,he2019AttGAN,liu2015deep}).}
\label{Two-stage of FAA}
\end{figure*}
\section{Facial Attribute Analysis Preliminaries}\label{section2}
Deep facial attribute analysis follows a general pipeline consisting of two stages: data preprocessing and model construction, as shown in Fig.~\ref{Two-stage of FAA}.

In this section, we first introduce two commonly used data preprocessing strategies for both FAE and FAM, including face detection and alignment, as well as data augmentation. Second, we introduce the general processes of model construction for deep FAE and FAM, respectively. Specifically, we provide the basics about feature extraction and attribute classification, which are two crucial steps when designing deep FAE models. For deep FAM methods, we review the underlying theories of backbone networks, i.e., VAEs and GANs, as well as their corresponding conditional versions.

\subsection{Data Preprocessing}\label{sub-section2-1}
\subsubsection{Face Detection and Alignment}
Before the databases with more facial attribute annotations were released, most of the attribute prediction methods \cite{zhang2014panda,kumar2008facetracer,gkioxari2015wholeandparts} took whole human images (faces and torsos) as inputs. Only several well-marked facial attributes could be estimated, i.e., \emph{smile}, \emph{gender}, and \emph{has glasses}. However, torso regions contain considerable face-irrelevant information, resulting in redundant computations. Hence, face detection and alignment become crucial steps to locate face areas for reducing the adverse effects of facial attribute-irrelevant areas.

For face detection, Ranjan et al. \cite{ranjan2017hyperface} first recognize the \emph{gender} attribute with a HyperFace detector that locates faces and landmarks, and then G{\"{u}}nther et al. \cite{GuntherRB17Affact} further extend this approach to predict 40 facial attributes simultaneously with the same HyperFace detector. In contrast, Kumar et al. \cite{kumar2008facetracer} use a poselet part detector \cite{Bourdev2009Poselets} to detect different parts corresponding to different poses, where the face is an important part of the whole person image. Compared with the poselet detector operated over conventional features, Gkioxari et al. \cite{gkioxari2015wholeandparts} propose a `deep' version of the poselet, which trains a sliding window detector operated on deep feature pyramids. Specifically, the deep poselet detector divides the human body into three parts (head, torso, and legs) and clusters fiducial key points of each part into many different poselets. However, because all existing face detectors are used to find rough facial parts, facial attributes in more subtle areas, such as \emph{eyebrows}, cannot be well predicted.

For facial alignment, well-aligned face databases with fiducial key points could alleviate the adverse effects of misalignment errors on both FAE and FAM when more specific facial regions of attributes can be located through these key points. The All-in-One Face algorithm \cite{ranjan2017allinone} can be utilized to obtain fiducial key points and full faces. Based on this algorithm, Mahbub et al. \cite{mahbub2018segment} divide a face into 14 segments related to different facial regions, and solve the problem of the attribute prediction in partial face images. Kumar et al. \cite{kumar2008facetracer} artificially divide a face into 10 functional parts including hair, forehead, eyebrows, eyes, nose, cheeks, upper lip, mouth, and chin. These facial areas are wide and robust enough to address discrepancies among individual faces, and the geometry characteristics shared by different faces can be well exploited.

Recently, researchers have tended to integrate face detection and alignment into the training process of facial attribute analysis. He et al. \cite{he2017jointlyFAEandDetec} take face detection as a special case of general semi-rigid object detection and design joint network architectures to ensure the performance improvement in both face detection and attribute estimation. More importantly, this approach can handle in-the-wild input images with complex illumination and occlusions, and no extra cropping and aligning operations are needed. Ding et al. \cite{ding2017deepCascadeFAE} propose a cascade network to locate face regions according to different attributes and perform FAE simultaneously with no need to align faces \cite{GuntherRB17Affact}. Li et al. \cite{li2018landmark} design an AFFAIR network for learning a hierarchy of spatial transformations and predicting facial attributes without landmarks. In summary, integrating face detection and alignment into the network training process is becoming a beneficial research trend. 
\subsubsection{Data Augmentation}
For most face processing tasks, data augmentation is a vital strategy for solving the problems of insufficient training data and overfitting in deep learning. Face attribute analysis is not an exception. By imposing perturbations and distortions on the input images, data can be extended to improve deep learning models.

G{\"{u}}nther et al. \cite{GuntherRB17Affact} propose an alignment-Free facial attribute classification technique (AFFACT) with data augmentation. More specifically, AFFACT leverages shifts, rotations, and scales of images to make facial attribute feature extraction more reliable in the training stage and the testing stage. In the training stage, face images are first rotated, scaled, cropped, and horizontally flipped with 50$\%$ probability with defined coordinates. Then, a Gaussian filter is applied to emulate smaller image resolutions and yield blurred upscaled images. In the testing stage, AFFACT first rescales the test images and then transforms these images into 10 crops, including a center one, four corners of the original images, and their horizontally flipped versions. Finally, AFFACT averages the scores from the deep network per attribute over the ten crops to make the final prediction. In addition to taking crops, AFFACT also uses all combinations of shifts, scales, and angles, as well as their mirrored versions. All these data augmentation schemes contribute to the progressive performance of deep FAE models.
\subsection{Basis of FAE Model Construction}
\subsubsection{Feature Extraction}
Deep convolutional neural networks (CNNs) play significant roles in learning discriminative representations and have achieved attractive performance in deep FAE. In general, arbitrary classical CNN networks, such as VGG \cite{parkhi2015deep} and ResNet \cite{he2016resnet}, can be used to extract deep facial attribute features. For example, Zhong et al. \cite{zhong2016face} directly apply FaceNet and VGG-16 networks to capture attribute features of face images.

Considering that the features at different levels of the network might have different effects on the performance of deep FAE methods, Zhong et al. \cite{zhong2016leveraging} take mid-level CNN features as an alternative to high-level features. The experiments demonstrate that even early convolution layers achieve comparable performance in most facial attributes with that of state-of-the-art methods, and mid-level representations can yield improved results over high-level abstract features. The reason for this superiority is that mid-level features can break the bounds of the inter-connections between convolutional and fully connected (FC) layers. Consequently, the CNN model can accept arbitrary receptive sizes for capturing rich information of face images.

In addition to using or combining classical deep networks, several methods design customized network architectures for learning discriminative features. Lu et al. \cite{lu2017fully} design an automatically constructed compact multi-task architecture, which starts with a thin multi-layer network and dynamically widens in a greedy manner. Belghazi et al. \cite{belghazi2018hierarchical} build a hierarchical generative model and a corresponding inference model through the adversarial learning paradigm.

\subsubsection{Attribute Classification}
Early methods learn feature representations with deep networks but make the prediction with traditional classifiers, such as support vector machines (SVMs) \cite{cortes1995support,bourdev2011describing}, decision trees \cite{luo2013sum-product}, and k-nearest neighbor (kNN) classifier \cite{huang2016learning,huang2019deep}. For example, Kumar et al. \cite{kumar2009attribute} train multiple SVMs \cite{cortes1995support} with radial basis function (RBF) kernels to predict multiple attributes, where each SVM corresponds to one facial attribute. Bourdev et al. \cite{bourdev2011describing} present a feedforward classification system with linear SVMs and classify attributes at the image patch level, the whole image level, and the semantic relationship level. Luo et al. \cite{luo2013sum-product} construct a sum-product decision tree network to yield facial attribute region locations and classification results simultaneously. Huang et al. \cite{huang2016learning,huang2019deep} adopt kNN algorithm to solve the class-imbalance attribute estimation problem.

In terms of the classifiers based on deep learning, several convolutional layers followed by FC layers constitute a deep attribute classifier, which can be attached to the end of deep feature extraction networks to make the prediction. Then, the specific loss function is used to measure the discrepancy between the outputs of FC layers and the ground truths for reducing classification errors. Below, we introduce two commonly used loss functions for deep FAE models.

The most prevalent loss function is the sigmoid cross-entropy loss, which makes a binary classification for each attribute \cite{hand2017attributes}. For example, Hand and Chellappa \cite{hand2017attributes} adopt the sigmoid cross-entropy loss to evaluate its network output and calculate the scores of all facial attribute. Besides, Rudd et al. \cite{rudd2016moon} consider multiple facial attribute classification as a regression issue to minimize the mean squared error (MSE) loss, i.e., the Euclidean loss, by mixing the errors of all attributes. In this way, multiple attribute labels can be obtained simultaneously via a single deep convolutional neural network (DCNN). In contrast, Rozsa et al. \cite{rozsa2016facial} also adopt the Euclidean loss but train a set of DCNNs, where each network predicts a facial attribute. Despite higher prediction accuracy that DCNNs achieve for facial attributes, they have the severe problem of high computation and memory costs.

To explore the effects of different loss functions on deep facial attribute classifiers, G{\"{u}}nther et al. \cite{GuntherRB17Affact} test and compare the Euclidean loss and the sigmoid cross-entropy loss. The experiments over the same network but different loss functions demonstrate that the two loss functions are capable of achieving comparable performance for attribute estimation. Therefore, future researchers can choose either of these loss functions according to their tasks with little performance change.
\subsection{Basis of FAM Model Construction}
\subsubsection{Variational autoencoder}
In general, a variational autoencoder (VAE) has two components: the generator, which samples the variables $x$ parameterized by $\theta$ with given latent variables $z$, i.e., ${p_\theta }(x|z)$; the encoder, which maps the variables $x$ to the latent variables $z$ that approximate a prior $p(z)$, i.e., ${q_\phi }(z|x)$ parameterized by $\phi $. The key of VAE is training to maximize the variational lower bound ${\cal L}_{VAE}$ \cite{huang2018introvae}:
\begin{equation}\label{vae}
\begin{aligned}
{\cal L}_{VAE} = {\mathbb{E}_{z \sim {q_\phi }(z|x)}}\log {p_\theta }\left( {x|z} \right)
-{D_{KL}}\left( {{q_\phi }\left( {z|x} \right)||p\left( z \right)} \right),
\end{aligned}
\end{equation}
where ${D_{KL}}$ denotes Kullback-Leibler divergence.

For the conditional version of VAE, given the attribute vector $y$ and latent representation $z$, it aims to build a model ${p_\theta }(x|y,z)$ for generating images $x$ that contain desired attributes, taking $y$ and $z$ as conditional variables. This image generation task follows a two-step process: the first step is randomly sampling the latent variables $z$ from the prior distribution $p(z)$, and the second step is generating an image according to the given conditional variables. Hence, the variational lower bound of conditional VAE can be written as \cite{yan2016attribute2image}
\begin{equation}\label{cvae}
\begin{aligned}
{{\cal L}_{CVAE}} &=\mathbb{E}_{z \sim {q_\phi }(z|x,y)}\log {p_\theta }\left( {x|y,z} \right) \\
&- {D_{KL}}\left( {{q_\phi }\left( {z|x,y} \right)||p\left( z \right)} \right),
\end{aligned}
\end{equation}
where ${q_\phi }(z|x,y)$ is the true posterior from the encoder.

\begin{table*}[t]
\renewcommand\arraystretch{1.05}
\begin{center}
\selftopcaption{An overview of facial attribute datasets. (Tra: Train, Tet: Test)}
\vspace{5pt}
\label{Datasets}
\resizebox{\textwidth}{!}{
\begin{tabular}{|c|c|c|c|c|c|}
  \hline
  Dataset & Resources & \tabincell{c}{Identities/\\ Samples} & \tabincell{c}{Number of\\attributes} & Protocol\\
  \hline
 FaceTracer \cite{kumar2008facetracer}\footnotemark[1]
 & Internet & 15,000 / 15,000 & 10 &\tabincell{c}{Tra: 80\% \\ Tet: 20\% \\ 5-fold cross-validation}\\
  \hline
 LFW \cite{huang2008labeled}\footnotemark[2]
 & Names and Faces \cite{miller2007names}
 & 5,749 / 13,233 & 65/73 &\tabincell{c}{Tra: 50\% (6,263) \\ Tet: 50\% (6,970)}\\
  \hline
 LFWA \cite{liu2015deep}\footnotemark[2]
 & LFW & 5,749 / 13,233 & 40 &\tabincell{c}{Tra: 50\%(6,263) \\ Tet: 50\%(6,970)}\\
  \hline
 PubFig \cite{kumar2009attribute}\footnotemark[3]
 & Internet & 200 / 58,797 & 73 &\tabincell{c}{Tra: 60 ids \\ Tet: 140 ids}\\
  \hline
 CelebA \cite{liu2015deep}\footnotemark[4]
 & Celeb-Faces & 10,177 / 202,599 & 40 &\tabincell{c}{Tra: 8000 ids (160,000)\\ Tet: 1000 ids (20,000)}\\
  \hline
 Berkeley Human Attributes \cite{bourdev2011describing}\footnotemark[5]
 & \tabincell{c}{H3D \cite{Bourdev2009Poselets} \\PASCAL VOC 2010 \cite{wang2016walk}} & - / 8,053 & 9 &\tabincell{c}{Tra: 2,003 \\ Tet: 4,022 \\Val: 2,010}\\
  \hline
 Attributes 25K \cite{zhang2014panda}
 & Facebook & 24,963 / 24,963 & 8 &\tabincell{c}{Tra: 8,737 ids \\ Tet: 7,489 ids\\Val: 8,437 ids}\\
  \hline
 Ego-Humans \cite{wang2016walk}
 & Videos & - / 2,714 & 17 &\tabincell{c}{Tra: 80\% \\ Tet: 20\%}\\
 \hline
 UMA-ADE \cite{hand2018doing}\footnotemark[6]
 & Image Research & - / 2,800 & 40 & All used for test\\
  \hline
 \tabincell{c}{YouTube Faces Dataset \cite{wolf2011face} \footnotemark[7] \\(with attribute labels)\cite{hand2018predicting}}  & \tabincell{c}{Videos \\from YouTube} &1,595 / 3,425 & 40 & \tabincell{c}{10-fold \\cross-validation}\\
  \hline
\end{tabular}}
\end{center}
\end{table*}

\subsubsection{Generative adversarial network}
A generative adversarial network (GAN) consists of two parts: the generator $G$ and the discriminator $D$, where $G$ attempts to synthesize data from a random vector $z$ obeying a prior noise distribution $z \sim p\left( z \right)$, and $D$ attempts to discriminate whether data is from the realistic data distribution or from $G$. Given data $x \sim {p_{data}}(x)$, $G$ and $D$ are trained in an adversarial manner with a min-max game as \cite{goodfellow2014gan}
\begin{equation}\label{gan}
\begin{aligned}
\mathop {\min }\limits_G \;\mathop {\max }\limits_D \;{{\cal L}_{GAN}} &= {\mathbb{E}_{x \sim {p_{data}}\left( x \right)}}\log \left( {D\left( x \right)} \right) \\ &+ {\mathbb{E}_{z \sim p\left( z \right)}}\log \left( {1 - D\left( {G\left( z \right)} \right)} \right).
\end{aligned}
\end{equation}

The conditional version of GAN is more frequently used by feeding the attribute vector $y$ into both $G$ and $D$ in different ways. Specifically, the attribute vector $y$ is concatenated with the prior input noise $p(z)$ in the generator. Meanwhile, it is taken as an input along with $x$ into a discriminative function. Therefore, the min-max game of conditional GAN is denoted as \cite{mirza2014conditional}
\begin{equation}\label{cgan}
\begin{aligned}
\mathop {\min }\limits_G \;\mathop {\max }\limits_D \;{{\cal L}_{CGAN}} &= {\mathbb{E}_{x \sim {p_{data}}\left( x \right)}}\log \left( {D\left( {x|y} \right)} \right) \\&+ {\mathbb{E}_{z \sim p\left( z \right)}}\log \left( {1 - D\left( {G\left( {z|y} \right)} \right)} \right).
\end{aligned}
\end{equation}

\section{Facial Attribute Analysis Datasets and Metrics}\label{section5}
\subsection{Facial Attribute Analysis Datasets}
We present an overview of publicly available facial attribute analysis datasets for both FAE and FAM, including data sources, sample sizes, and test protocols. More details of these datasets are listed in Table~\ref{Datasets}.

\textbf{FaceTracer dataset} is an extensive collection of real-world face images collected from the internet. There are 15,000 faces with fiducial key points and 10 groups of attributes, where 7 groups of facial attributes are composed of 19 attribute values, and the remaining 3 groups denote the quality of images and the environment. This dataset provides the URLs of each image for considering privacy and copyright issues. In addition, FaceTracer takes 80\% of the labeled data as training data, and the remaining 20\% as testing data with 5-fold cross-validation.

\textbf{The Labeled Faces in the Wild (LFW) dataset} consists of 13,233 images of cropped, centered frontal faces derived from T. Berg et al. \cite{miller2007names}. This dataset is collected from 5,749 people using online news sources, and there are 1,680 people that have two or more images. Kumar et al. \cite{kumar2009attribute} first collect 65 attribute labels through Amazon Mechanical Turk (AMT) \cite{AmazonMechanicalTurk} and then expand to 73 attributes \cite{kumar2011visual_attributes}. We denote them as LFW-65 and LFW-73 in Table~\ref{Facial Attributes}. Liu et al. \cite{liu2015deep} extract 40 attribute labels automatically by binarizing corresponding values of labels in LFW dataset, instead of labeling by manual. Moreover, they annotate 5 fiducial key points, leading to \textbf{LFWA dataset}, which is partitioned into half for training (6,263 images) and the remains for testing.

\textbf{PubFig dataset} is a large, real-world face dataset containing 58,797 images of 200 people collected from the internet under uncontrolled situations. Thus, this dataset covers considerable variations in poses, lights, expressions, and scenes. PubFig dataset labels 73 facial attributes, as many as LFW-73, but it includes fewer individuals. Besides, this dataset divides the development set and the evaluation set, containing 60 identity images and 140 identities, respectively.
\footnotetext[1]{\url{www.cs.columbia.edu/CAVE/databases/facetracer/}}
\footnotetext[2]{\url{http://vis-www.cs.umass.edu/lfw/}}
\footnotetext[3]{\url{http://www.cs.columbia.edu/CAVE/databases/pubfig/download/}}
\footnotetext[4]{\url{http://mmlab.ie.cuhk.edu.hk/projects/CelebA.html}}
\footnotetext[5]{\url{https://www2.eecs.berkeley.edu/Research/Projects/CS/vision/shape/poselets/}}
\footnotetext[6]{\url{https://www.cs.umd.edu/~emhand/research.html}}
\footnotetext[7]{\url{https://www.cs.tau.ac.il/~wolf/ytfaces/}}

\textbf{Celeb-Faces Attributes (CelebA) dataset} is constructed by labeling images selected from Celeb-Faces \cite{sun2014deep}, which is a large-scale face attribute dataset covering large pose variations and background clutter. There are 10,177 identities, 202,599 face images with 5 landmark locations, and 40 binary attribute annotations per image. In the experiment, CelebA is partitioned into three parts: images of the first 8,000 identities (with 160,000 images) for training, images of another 1,000 identities (with 20,000 images) for validation and the remains for testing.

\textbf{Berkeley Human Attributes dataset} is collected from H3D \cite{Bourdev2009Poselets} dataset and PASCAL VOC 2010 \cite{wang2016walk} training and validation datasets, containing 8,053 images centered on full bodies of persons. There are wide variations in poses, viewpoints, and occlusions. Thus, many existing methods that work on front faces do not perform well on this dataset. AMT is also used to provide labels for all 9 attributes by 5 independent annotators. The dataset partitions 2,003 images for training, 2,010 for validation and 4,022 for testing.

\textbf{Attribute 25K dataset} is collected from Facebook, which contains 24,963 people split into 8,737 training, 8,737 validation and 7,489 test examples. Since the images have large variations in viewpoints, poses and occlusions, not every attribute can be inferred from every image. For instance, we cannot label the \emph{wearing hat} attribute when the head of the person is not visible.

\textbf{Ego-Humans dataset} draws images from videos that track casual walkers with the OpenCV frontal face detector and facial landmark tracking in New York City over two months. What makes it different from other datasets is that it covers the location and weather information through clustering GPS coordinates. Moreover, nearly five million face pairs along with their same or not same labels are extracted under the constraints of temporal information and geolocations. Wang et al. \cite{wang2016walk} manually annotate 2,714 images with 17 facial attributes randomly selected from these five million images. For the testing protocol, 80\% images are selected randomly for training and the remaining for testing.

\textbf{University of Maryland Attribute Evaluation Dataset (UMA-AED)} comes from image searches taking 40 attributes as search terms and the HyperFace as face detector \cite{ranjan2017hyperface}. UMD-AED serves as an evaluation dataset and contributes to class-imbalance learning in deep facial attribute estimation. It is composed of 2,800 face images labeled with a subset of 40 attributes from CelebA and LFWA. Each attribute has 50 positive and 50 negative samples, which means that not every attribute is tagged in each image. In addition, compared with CelebA containing mostly frontal, high-quality, and posed images, UMD-AED comprises a large number of variations, e.g., distinct image quality, varying lights and poses, wide age ranges, and different skin tones. UMD-AED offers a much more unbiased metric for real-world data, and it can be used to evaluate whether the attribute estimation models have learned discriminative feature representations.

\textbf{YouTube Faces Dataset (with attribute labels)} Original YouTube Faces Dataset contains 3,245 videos from 1,595 celebrities with 620,000 frame images \cite{wolf2011face} for face verification. Hand et al. \cite{hand2018predicting} further extend it for the video-based facial attribute prediction issue. They label 40 attributes from CelebA in the first of four frames from every video, where the remaining three frames without attribute labels come from one third, two-thirds, and the last of the way per video, respectively. As a result, this dataset makes it possible for exploring deep FAE methods merely with weakly labels. Ten-fold cross-validation is adopted for the protocol. Then, all the testing experiments need to be conducted on the labeled frames of the testing splits with the average of all 10 splits.

To provide a more comprehensive overview of all existing attribute labels, we list all the labels in LFW dataset with the maximum number of attributes in Table~\ref{Facial Attributes}. Different facial attribute datasets contain different subsets of these attribute annotations for deep FAE and FAM. Note that in Table~\ref{Facial Attributes}, `Common' denotes the attributes shared by all variants of LFW, a total of 34 categories. CelebA and LFWA have more 6 attributes than `Common' has and share a total of 40 common attributes. Besides, these 6 attributes, together with the underlined \emph{flushed face} and \emph{brown eyes}, are added to LFW-65 to constitute LFW-73. To date, the most popular and commonly used datasets in both FAE and FAM are CelebA and LFW (LFWA).
\begin{table*}[t]
\renewcommand\arraystretch{1.4}
\begin{center}
\caption{An overview of facial attributes}
\label{Facial Attributes}
\resizebox{\textwidth}{!}{
\begin{tabular}{|c|ccccccccc|}
  \hline
  Dataset & \multicolumn{9}{c|}{Attributes}\\
  \hline
  \multirow{11}{*}{LFW}
                & \multicolumn{2}{c|}{\multirow{5}{*}{Common}}
                & Arched Eyebrows & Attractive & Bags under eyes & Bald & Bangs & Big nose & Black hair \\
                &\multicolumn{2}{c|}{}& Blond hair  & Blurry & Brown hair & Bushy eyebrows & Chubby & Double chin & Eyeglasses\\
                &\multicolumn{2}{c|}{}& Goatee & Gray hair & High cheekbones & Male & Mouth slightly open & Mustache & Narrow eyes \\
                &\multicolumn{2}{c|}{}& No beard & Oval face & Pale skin & Pointy nose & Receding hairline & Rosy cheeks & Sideburns \\
                &\multicolumn{2}{c|}{}& Smiling & Straight hair & Wavy hair & Wearing hat& Wearing lipstick & Young &\\\cline{2-10}

                & \multicolumn{1}{c|}{\multirow{6}{*}{LFW-73}}
                & \multicolumn{1}{c|}{LFWA/CelebA}
                & Big lips  & Heavy makeup & Wearing earrings & Wearing necklace & Wearing necktie & 5'o clock shadow &\\\cline{3-10}

                & \multicolumn{1}{c|}{}&\multicolumn{1}{c|}{\multirow{5}{*}{LFW-65}}
                & Asian & Baby & Black & Child & Color photo & Curly hair & Environment \\
                &\multicolumn{1}{c|}{}& \multicolumn{1}{c|}{}& Eyes open & Flash & Frowning &Fully visible forehead & Harsh lighting & Indian & Middle aged \\
                &\multicolumn{1}{c|}{}& \multicolumn{1}{c|}{}& Mouth wide open & Mouth closed & No eyewear  & Obstructed forehead & Posed photo & Round jaw & Round face \\
                &\multicolumn{1}{c|}{}& \multicolumn{1}{c|}{}& Semi obscured forehead & Senior & Shiny skin & Soft lighting & Square face & Strong nose mouth lines& Sunglasses \\
  &\multicolumn{1}{c|}{}& \multicolumn{1}{c|}{}& Teeth not visible  & Teeth visible & White & & & (\underline{\emph{Flushed face}} & \underline{\emph{Brown eyes}})\\
\hline
\end{tabular}}
\end{center}
\end{table*}
\subsection{Facial Attribute Analysis Metrics}
\subsubsection{Facial Attribute Estimation Metrics}\label{Section_3.2.1}
Below, we list the frequently used metrics for FAE algorithms and provide detailed descriptions of these metrics in terms of definitions and formulas.
\begin{itemize}
    \item \textbf{Accuracy and Error Rate (Acc and ER)}
\end{itemize}

The classification accuracy and the error rate are the most commonly used measures for evaluating classification tasks. Facial attribute estimation is not an exception, and its accuracy can be defined as \cite{rudd2016moon}
\begin{equation}\label{ACC}
Accuracy = \left( {\left( {{t_p} + {t_n}} \right)/\left( {{N_p} + {N_n}} \right)} \right).
\end{equation}
where $N_p$ and $N_n$ denote the numbers of positive and negative samples, respectively, and $t_p$ and $t_p$ denote the numbers of true positives and true negatives \cite{huang2016learning}. Meanwhile, the error rate can be defined as
\begin{equation}\label{ErrorRate}
Error\;Rate = 1 - Accuracy.
\end{equation}
\begin{itemize}
    \item \textbf{Balanced Accuracy and Error Rate (BAcc and BER)}
\end{itemize}

When dealing with class-imbalance data, the traditional classification accuracy is not befitting due to the bias of the majority class. Hence, a balanced classification accuracy is defined as \cite{rudd2016moon}
\begin{equation}\label{Balanced ACC}
Balanced\;Accuracy = \frac{1}{2}\left( {{t_p}/{N_p} + {t_n}/{N_n}} \right).
\end{equation}
Similarly, the balanced error rate can be defined as $Balanced\;Error\;Rate = 1 - Balanced\;Accuracy$. When addressing the imbalance issue from the perspective of source and target distributions \cite{rudd2016moon}, the balanced error rate is defined as
\begin{equation}\label{Balanced Error Rate}
Balanced\;Error\;Rat{e^*} = \left( {{T^ + }\left( {{t_p}/{N_p}} \right) + {T^ - }\left( {{t_n}/{N_n}} \right)} \right),
\end{equation}
where $T^+$ and $T^-$ denote the target domain distributions of positive and negative examples, respectively. The superscript $*$ is used to indicate the balanced version of error rate. Besides, more details of the class-imbalance issue are introduced in Section~\ref{section6_1}.
\begin{itemize}
    \item \textbf{mean Average Precision (mAP)}
\end{itemize}

As there is more than one label in multi-label image classification, the mean Average Precision (mAP) becomes a prevalent metric \cite{yue2007support,philbin2007object}, which computes the average of the precision$-$recall curve from the recall 0 to recall 1. Moreover, mAP is the mean of Average Precision (AP) for a set of categories, while AP is the more general version that combines the recall and precision to yield prediction results for a single class.
\subsubsection{Facial Attribute Manipulation Metrics}
There are two types of measurements in deep FAM: qualitative metrics and quantitative metrics, where the former evaluates the performance of generated images through statistical surveys, and the latter measures the preservation degree of the face detail related information after attribute manipulation. We provide more detailed descriptions of these two types of metrics below.

\begin{itemize}
    \item \textbf{Qualitative Metrics}
\end{itemize}

\textbf{Statistical survey} is the most intuitive way to qualitatively evaluate the quality of generated images in most generative tasks. By establishing specific rules in advance, subjects vote for generated images with appealing visual fidelity, and then, researchers draw conclusions according to the statistical analysis of votes. For example, Choi et al. \cite{choi2017stargan} quantitatively evaluate the performance of generated images in a survey format via AMT \cite{AmazonMechanicalTurk}. Given an input image, the workers are required to select the best generated images according to instructions based on perceptual realism, quality of manipulation in attributes, and preservation of original identities. Each worker is asked a set number of questions for validating human effort.

Zhang et al. \cite{zhang2017age} conduct a statistical survey that asks volunteers to choose the better result from their proposed CAAE or existing works. Sun et al. \cite{sun2018mask} instruct volunteers to rank several deep FAM approaches based on perceptual realism, quality of transferred attributes, and preservation of personal features. Then, they calculate the average rank (between 1 and 7) of each approach. Lample et al. \cite{lample2017fader} perform a quantitative evaluation on two different aspects: the naturalness measuring the quality of generated images, and the accuracy measuring the degree of swapping an attribute reflected in the generation.

\begin{itemize}
    \item \textbf{Quantitative Metrics}
\end{itemize}

\textbf{Distribution difference measure} calculates the differences between real images and generated face images. Xiao et al. \cite{xiao2018elegant} achieve this goal by the Fr\'echet inception distance \cite{heusel2017gans} (FID) with the means and covariance matrices of two distributions before and after editing facial attributes. Wang et al. \cite{wang2018weakly} compute the peak signal to noise ratio (PSNR) to measure the pixel-level differences. They also calculate the structure similarity index (SSIM) and its multi-scale version MS-SSIM \cite{wang2004assess} to estimate the structure distortion and the identity distance. All these measurements contribute to evaluating the high-level similarity of two face images. In addition, He et al. \cite{he2019AttGAN} use an Inception-ResNet \cite{szegedy2017inception} to train a face recognizer for measuring the identity preservation ability with rank-1 recognition accuracy. Therefore, face identity preservation is becoming a promising metric because it can indicate whether models have excellent performance in preserving facial details outside of manipulated attributes.
\begin{figure*}[t]
\centering
{\includegraphics[width=0.7\linewidth]{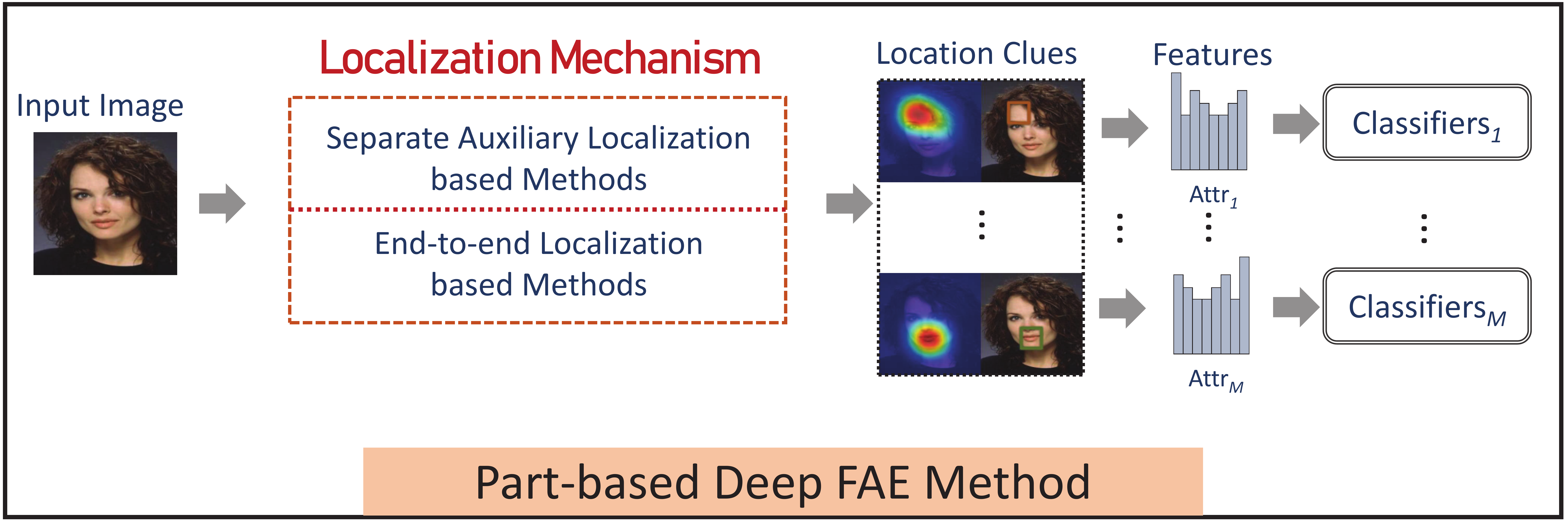}}
\caption{The illustration of deep part-based FAE methods (Images are from \cite{ding2017deepCascadeFAE}).}
\label{Partbased_FAE}
\end{figure*}

\textbf{Facial landmark detection gain} uses the accuracy gain of landmark detection before and after attribute editing to evaluate the quality of synthesized images. For example, He et al. \cite{he2016dual} adopt an ERT method \cite{kazemi2014one}, which is a landmark detection algorithm trained on 300-W dataset \cite{sagonas2013semi}. During testing, they divide the test sets into three components: the first containing images with the positive attribute labels, the second containing images with the negative labels, and the last containing the manipulated images from the first part. Then, the average normalized distance error is computed to evaluate the discrepancy of landmarks between the generated images and the ground truths.

\textbf{Facial attribute estimation} constructs additional attribute prediction networks to measure the performance of FAM according to the classification accuracy. Perarnau et al. \cite{perarnau2016invertible} first design an Anet to predict facial attributes on the manipulated face images. If the outputs of the Anet are closer to the desired attribute labels, the generator can be considered to have satisfactory generation performance. Larsen et al. \cite{larsen2016autoencoding} train a regressor attribute prediction network to calculate the attribute similarity between the conditional attributes and generated attributes. Note that FAE models used for the evaluation are independent of FAM's training processes, which means that they have to be trained well in advance and have base accuracy performance over all facial attributes.

\section{State-of-the-art Facial Attribute Estimation Methods} \label{section3}
Generally, state-of-the-art deep FAE methods can be divided into two main categories: part-based methods and holistic methods. In this section, we provide detailed introductions to these two types of methods in terms of algorithms, performance, as well as their respective advantages and disadvantages. The overview is provided in Table~\ref{Deep FAE}.
\subsection{Part-based Deep FAE Methods}
As shown in Fig~\ref{Partbased_FAE}, part-based deep FAE methods first locate the areas where facial attributes exist through localization mechanisms. Then, features corresponding to different attributes on each highlighted position can be extracted and further predicted with multiple attribute classifiers. Hence, the key of part-based methods lies in the localization mechanism. In light of this point, part-based deep FAE methods can be further divided into two subgroups: separate auxiliary localization based methods and end-to-end localization based methods. Corresponding details are provided as follows.
\subsubsection{Separate Auxiliary Localization based Methods}
Since facial attributes describe subtle details of face representations based on human vision, locating the positions of facial attributes enforces subsequent feature extractors and attribute classifiers to focus more on attribute-relevant regions. The most intuitive approach is to take existing face part detectors as auxiliaries.

Poselet \cite{Bourdev2009Poselets,bourdev2011describing} is a valid part detector that describes a part of the human pose under a given viewpoint. Because these parts include evidences from different areas of the body at different scales, complementary information can be learned to benefit attribute prediction. Typically, given a whole person image, poselet detector \cite{zhang2014panda} is first used to decompose an image into several image patches, named poselets, under various viewpoints and poses. Then, a PANDA network is proposed to train a set of CNNs for each poselet and the whole image. Then, the features from all these poselets are concatenated to yield final feature representations. Finally, PANDA branches out multiple binary classifiers where each recognizes an attribute by the binary classification. Based on PANDA, Gkioxari et al. \cite{gkioxari2015wholeandparts} introduce a deep version of the Poselet detector and build a feature pyramid, where each level computes a prediction score for the corresponding attribute.

However, the poselet detector only discovers coarse body parts and cannot explore subtle local details of face images. Considering that the probability of an attribute appearing in a face image is not uniformed in the spatial domain, Kalayeh et al. \cite{kalayeh2017improving} propose employing semantic segmentation as a separate auxiliary localization scheme. They exploit the location cues obtained by semantic segmentation to guide the attention of attribute prediction to the naturally occurring areas of attributes. Specifically, a semantic segmentation network is first designed in an encoder-decoder paradigm and trained over Helen face dataset \cite{le2012interactive}. During this process, the semantic face parsing \cite{smith2013exemplar,lu2018conditional} is performed as an additional task to learn detailed pixel-level location information. After discovering the location cues, the semantic segmentation based pooling (SSP) and gating (SSG) mechanisms are presented to integrate the location information into the attribute estimation. SSP decomposes the activations of the last convolutional layer into different semantic regions and then aggregates those regions that only reside in the same area. Meanwhile, SSG gates the output activations between the convolutional layers and the batch normalization (BN) operation to control the activations of neurons from different semantic regions.

In contrast, Mahbub et al. \cite{mahbub2018segment} utilize key points to segment faces into several image patches, which is a more straightforward way compared with semantic segmentation. Then, these segments are fed into a set of facial segment networks to extract corresponding feature representations and learn prediction scores, where the whole face image is fed into a full-face network. A global predictor network fuses the features from these segments, and two committee machines merge their scores for the final prediction.

Compared with the above methods that search for location clues of attributes directly, He et al. \cite{he2018harnessing} resort to synthesized abstraction facial images that contain local facial parts and texture information to achieve the same goal indirectly. A designed GAN is used to generate facial abstraction images before inputting them into a dual-path facial attribute recognition network, where the real original images are together fed into this recognition network. The dual-path network propagates the feature maps from the abstraction sub-network to the real original image sub-network and concatenates the two types of features for the final prediction. Despite the abundant location and textual information that is contained in generated facial abstraction images, the quality of these images may have a significant impact on performance, especially when some attribute related information is lost in image abstraction.

Note that all the separated auxiliary localization based deep FAE methods share a common drawback: relying too much on accurate facial landmark localization, face detection, facial semantic segmentation, face parsing, and facial partition schemes. If these localization strategies are imprecise or landmark annotations are unavailable, the performance of the subsequent attribute estimation task would be significantly affected.

\subsubsection{End-to-end Localization based Methods}
Compared with the separate auxiliary localization based methods that locate attribute regions and make the attribute prediction separately and independently, end-to-end localization based methods jointly exploit location cues where facial attributes appear and predict their presence in a unified framework.

Liu et al. \cite{liu2015deep} first propose a cascaded deep learning framework for joint face localization and attribute prediction. Specifically, the cascaded CNN is made up of an LNet and an ANet, where the LNet locates the entire face region and the ANet extracts the high-level face representation from the located area. LNet is first pretrained by classifying massive general object categories to ensure excellent generalization capability, and then it is fine-tuned using the image-level attribute tags of training images to learn features for face localization in a weakly supervised manner. Note that the main difference between LNet and separated auxiliary localization based methods is LNet does not require face bounding boxes or landmark annotations. Meanwhile, ANet is first pretrained by classifying massive face identities to handle the complex variations in unconstrained face images, and then it is fine-tuned to extract discriminative facial attribute representations. Furthermore, rather than extracting features patch-by-patch, ANet introduces an interweaved operation with locally shared filters to extract multiple feature vectors in a one-pass feed-forward process. Finally, SVMs are trained over these features to estimate attribute values per attribute, and the terminal prediction is made by averaging all these values for addressing the small misalignment of face localization. The cascaded LNet and ANet framework shows the benefit of pretraining with massive object categories and massive identities in enhancing the feature representation learning. With such customized pretraining schemes and cascaded network architecture, this method exhibits outstanding robustness to backgrounds and face variations.

However, coarse entire face regions discovered by LNet cannot be used to explore more local attribute details. Hence, Ding et al. \cite{ding2017deepCascadeFAE} propose a cascade network to jointly locate facial attribute-relevant regions and perform attribute classification. Specifically, they first design a face region localization network (FRL) that builds a branch for each attribute to automatically detect a corresponding relevant region. Then, the following parts and whole (PaW) attribute classification network selectively leverages information from all the attribute-relevant regions for the final estimation. Moreover, in terms of the attribute classification, Ding et al. define two FC layers: the region switch layer (RSL) and the attribute relation layer (ARL). The former selects the relevant prediction sub-network and the latter models attribute relationships. In summary, the cascaded FRL and PaW model not only discovers semantic attribute regions but also explores rich relationships among facial attributes. Besides, since this model automatically detects face regions, it can achieve outstanding performance on unaligned datasets without any pre-alignment step.
\begin{figure*}[t]
\centering
{\includegraphics[width=0.7\linewidth]{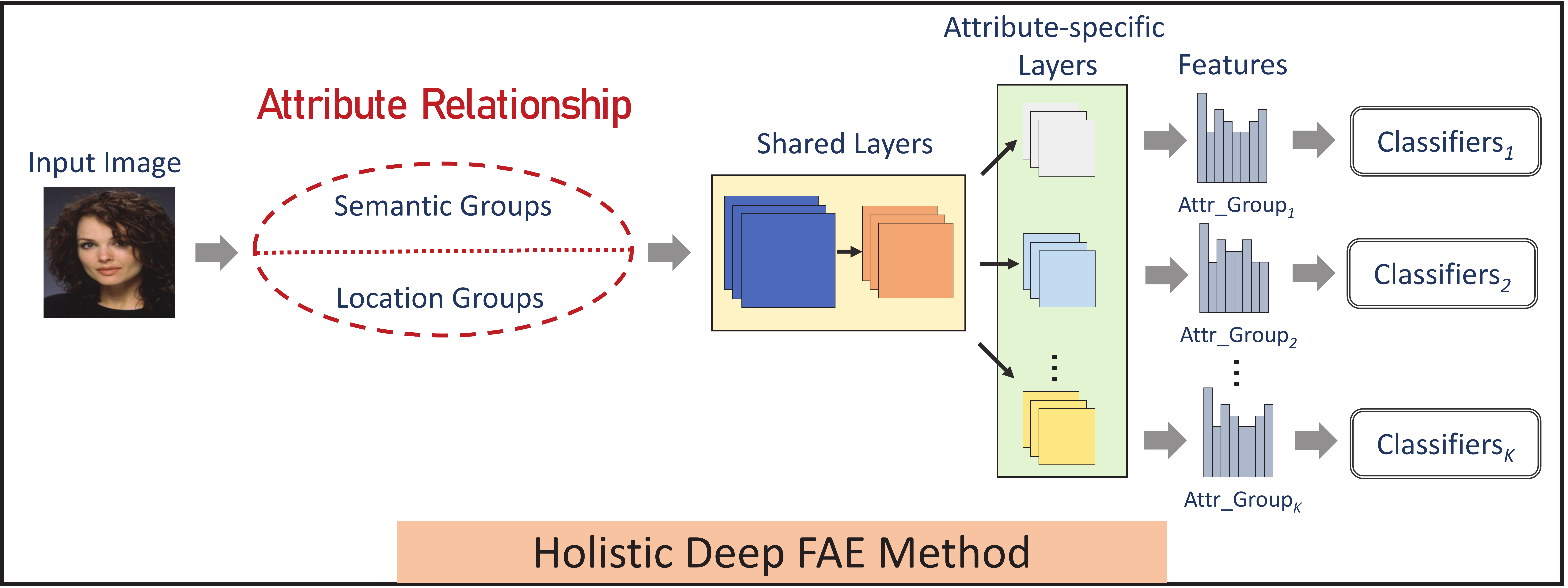}}
\caption{The illustration of deep holistic FAE methods (Face image comes from \cite{ding2017deepCascadeFAE}).}
\label{Holistic_FAE}
\end{figure*}

Note that FRL-PaW method learns a location for each attribute, which makes the training process redundant and time-consuming. This is because several facial attributes generally exist in the same area. However, to the best of our knowledge, there is currently no specific solution for tackling this issue. We expect that future research would reduce computation costs; meanwhile, make the prediction according to attribute locations as accurately as possible.

In summary, part-based deep FAE methods first locate the positions where facial attributes appear. Two strategies can be adopted: separate auxiliary localization and end-to-end localization. The former leverages existing part detectors or auxiliary localization-related algorithms, and the latter jointly exploits the locations in which facial attributes exist and predicts their presences. Compared with the separate auxiliary localization based methods operating separately and independently, end-to-end localization based methods locate and predict in a unified framework. After obtaining the location clues, features corresponding to certain attribute areas can be extracted and further be fed into attribute classifiers to make the estimation. Recently, researchers are currently more inclined to shift their focus on holistic FAE algorithms when the part-based counterparts are generally distracted and affected by attribute localization mechanisms.

\subsection{Holistic Deep FAE Methods}
In contrast to part-based FAE approaches that detect and utilize facial components, holistic deep FAE methods focus more on exploring the attribute relationships and extracting features from entire face images rather than facial parts. A schematic diagram of holistic FAE models is provided in Fig.~\ref{Holistic_FAE}.

As shown in Fig.~\ref{Holistic_FAE}, the key to modeling attribute relationships is learning common features at low-level shared layers and capturing attribute-specific features at high-level separated layers. Each separated layer corresponds to an attribute group. In general, these attribute groups are obtained manually according to semantics or attribute locations. By assigning different shared layers and attribute-specific layers, complementary information among multiple attributes can be discovered such that more discriminative features can be learned for the following attribute classifiers.

In general, there are two crucial issues that holistic deep FAE methods need to address when designing network architectures: (1) how to properly assign shared information and attribute-specific information at different layers of networks, and (2) how to explore relationships among facial attributes for learning more discriminative features. Taking these two problems as the main focus, we provide a brief review of holistic FAE methods in the following parts.

To the best of our knowledge, MOON \cite{rudd2016moon} is one of the earliest holistic FAE methods with the multi-task framework. It has a mixed objective optimization network that learns multiple attribute labels simultaneously via a single DCNN. MOON takes deep FAE as a regression problem for the first time and adopts a 16-layer VGG network as the backbone network, in which abstract high-level features are shared before the last FC layer. Multiple prediction scores are calculated with the MSE loss to reduce the regression error. Similarly, Zhong et al. \cite{zhong2016leveraging} replace the high-level CNN features in MOON with mid-level features to identify the best representation for each attribute.

\begin{table*}[t]
\renewcommand\arraystretch{1.4}
\begin{center}
\caption{State-of-the-art Deep Facial Attribute Estimation Approaches.}\label{Deep FAE}
\resizebox{\textwidth}{!}{
\begin{tabular}{|c|c|c|c|}
  \hline
  Approaches & Algorithms & Network Architectures & \tabincell{c}{Datasets \\ Metrics and Performance} \\\hline\hline
  \multicolumn{4}{|c|}{\emph{PART-BASED METHODS}} \\\hline
  PANDA \cite{zhang2014panda}(CVPR2014) & \tabincell{c}{Using Part-based Pose Aligned Networks \\for learning features related to poses and\\ Linear SVM classifiers for attribute estimation} & PANDA & \tabincell{c}{Berkeley Human Attributes Dataset (mAP:78.98\%)\\ Attribute 25K Datasets (mAP:70.74\%)\\LFW-gender (mAP:99.54\%)}\\
  \hline
  Gkioxari et al. \cite{gkioxari2015wholeandparts}(ICCV2015) & \tabincell{c}{Using a deep version of poselets \\ and capturing parts of the human body \\for tasks of action and attribute classification} & \tabincell{c}{A 5-layer CNN feature pyramid \\and a  pyramid of part scores} & Berkeley Human Attributes Dataset (mAP:89.5\%) \\
  \hline
  LNet$+$ANet \cite{liu2015deep}(ICCV2015) & \tabincell{c}{Cascading LNet CNN for localization \\and ANet for feature extraction} & LNet$+$ANet & \tabincell{c}{CelebA (Acc:87\%)\\LFWA (Acc:84\%)} \\
  \hline
  Off-the-shelf CNN \cite{zhong2016face}(ICB2016)&\tabincell{c}{Training off-the-shelf architectures for \\ face recognition to construct facial representations} & Off-the-shelf & \tabincell{c}{CelebA (Acc:86.6\%)\\LFWA (Acc:84.7\%)} \\
  \hline
  Singh et al. \cite{singh2016end}(ECCV2016)&\tabincell{c}{Using Spatial Transformer Network (STN) and \\Ranker Network (RN) to jointly learn features,\\ localization and ranker of attributes} & STN and RN & LFW-10attr (Rank Accuracy:86.91\%)\\
  \hline
  SSP$+$SSG \cite{kalayeh2017improving}(CVPR2017)&\tabincell{c}{Using semantic segmentation guiding the attention \\of the attribute prediction to the regions \\where different attributes naturally show up}& \tabincell{c}{Semantic Segmentation based \\
  Pooling (SSP) and Gating (SSG)}& \tabincell{c}{CelebA (ER:8.20\%)\\(mAP:81.45\%)\\(BAcc:88.24\%)}\\
  \hline
  FRL-PaW \cite{ding2017deepCascadeFAE}(AAAI2018)&\tabincell{c}{Simultaneously learning to localize face regions \\ specific to attributes and performs attribute classification \\without alignment in a cascade network}& \tabincell{c}{Facial region localization (FRL) network\\Parts and Whole (PaW) classification network}& Unaligned CelebA (Acc:91.23\%)\\
  \hline
  SPLITFACE \cite{mahbub2018segment}(IEEE TAC2018)&\tabincell{c}{Using facial segmentation for attribute \\detection in partially occluded faces}&\tabincell{c}{Segmentwise, Partial,Localized \\Inference in Training Facial Attribute \\Classification Ensembles (SPLITFACE) Network}& CelebA (Acc:90.61\%)\\
  \hline
  FMTNet \cite{zhuang2018DeepTranfer}(PR2018)&\tabincell{c}{Constructing three sub-networks \\for attribute transfer learning}& \tabincell{c}{the Face detection Network (FNet)\\the Multi-label learning Network (MNet)\\the Transfer learning Network (TNet)}&\tabincell{c}{CelebA (Acc:91.66\%)\\LFWA (Acc:84.34\%)}\\
  \hline
  He et al. \cite{he2018harnessing}(IJCAI2018)&\tabincell{c}{Generating abstraction images by GAN as \\complementary features and used for facial parts localization}& \tabincell{c}{GAN and a dual-path facial \\attribute recognition network}&\tabincell{c}{CelebA (Acc:91.81\%)\\LFWA (Acc:85.2\%)}\\
  \hline
  AFFAIR \cite{li2018landmark}(IEEE TIP2018)&\tabincell{c}{Learning a hierarchy of spatial transformations for \\facial attribute prediction with no landmark}&\tabincell{c}{lAndmark Free Face AttrIbute \\pRediction (AFFAIR) Network}&\tabincell{c}{CelebA (mAP:79.63\%/Acc:91.45\%)\\LFWA (mAP:83.01\%/Acc:86.13\%)\\MTFL (Acc:86.55\%)}\\
  \hline\hline
  \multicolumn{4}{|c|}{\emph{HOLISTIC METHODS}} \\\hline
  Wang et al. \cite{wang2016walk}(CVPR2016)&\tabincell{c}{Employing a Siamese structure, \\embedding location and weather contextual information} & Siamese & \tabincell{c}{CelebA (Acc:88\%)\\LFWA (Acc:87\%)\\Ego-Humans Dataset(Acc:87\%)}\\
  \hline
  MOON \cite{rudd2016moon}(ICCV2016) &\tabincell{c}{Treating attribute classification as a \\regression task and solving domain adaptive problem} & \tabincell{c}{Mixed-Objective Optimization Network \\(MOON, VGG16)}& \tabincell{c}{CelebA(ER:9.06\%) \\CelebAB (BER:13.67\%)}\\
  \hline
  LMLE \cite{huang2016learning}(CVPR2016)& \tabincell{c}{Using a Large Margin Local Embedding (LMLE) Method for \\ large-scale imbalanced facial attribute classification} & VGG-6 Quintuplet CNN & \tabincell{c}{CelebA(BAcc:84.25\%)}\\
  \hline
  Zhong et al. \cite{zhong2016leveraging}(ICIP2016)& \tabincell{c}{Studying the effect of mid-level \\ CNN features for attribute prediction} & FaceNet NN.1 \cite{schroff2015facenet} &\tabincell{c}{CelebA (Acc:89.8\%)\\LFWA (Acc:85.9\%)}\\
  \hline
  CRL \cite{dong2017CRL}(ICCV2017) &\tabincell{c}{Combining batch-wise incremental hard mining \\for class-imbalance learning with the \\Class Rectification Loss (CRL) regularizing algorithm}& 5-layer DeepID2 \cite{sun2014deep} CNN& \tabincell{c}{CelebA(BAcc:86\%)}\\
  \hline
  AFFACT \cite{GuntherRB17Affact}(IJCB2017)&\tabincell{c}{Introducing the Alignment-Free Facial Attribute Classification Technique} & AFFACT Network (ResNet)& CelebA (ER:8.03\%)\\
  \hline
  MCNN$+$AUX \cite{hand2017attributes}(AAAI2017)&\tabincell{c}{Considering attribute relationships \\and constructing a Multi-task deep CNN (MCNN)\\ with an Auxiliary Network (AUX) for performance improvement}& MCNN$+$AUX & \tabincell{c}{CelebA (Acc:91.22\%)\\LFWA (Acc:86.31\%)}\\
  \hline
  DMTL \cite{han2017heterogeneous}(IEEE TPAMI2017)&\tabincell{c}{Introducing Deep multi-task feature learning (DMTL) for \\joint estimation of multiple heterogeneous attributes}& DMTL (AlexNet)& \tabincell{c}{CelebA (Acc:93\%)\\LFWA (Acc:86\%)}\\
  \hline
  Lu et al.\cite{lu2017fully}(CVPR2017) & \tabincell{c}{Automatically designing compact multi-task deep network} & Automatic top-down layer-wise widening & \tabincell{c}{CelebA \\(Acc:91.02\%)\\(Top-10 Recall:71.38\%)}\\
  \hline
  AttCNN \cite{hand2018doing}(AAAI2018)&\tabincell{c}{Selectively learning with domain adaptive batch resample \\ methods for multi-label attribute prediction}& AttCNN Network &\tabincell{c}{CelebA (Acc:85.05\%)\\LFWA (Acc:73.03\%)\\UMD-AED (Acc:71.11\%)}\\
  \hline
  R-Codean \cite{sethi2018residual}(PRLetters2018)& \tabincell{c}{Incorporating a Cosine similarity based\\ loss function into the Euclidean distance \\for constructing an R-Codean autoencoder}& Residual Codean Autoencoder &\tabincell{c}{CelebA (Acc:90.14\%)\\LFWA (Acc:84.90\%)}\\
  \hline
  PS-MCNN \cite{cao2018partially}(CVPR2018)&\tabincell{c}{Considering the identity information and \\attribute relationships simultaneously and constructing \\a Partially Shared Multi-task Convolutional Neural Network} & PS-MCNN &\tabincell{c}{CelebA (ER:7.02\%)\\LFWA (ER:12.64\%)}\\
  \hline
\end{tabular}}
\end{center}
\end{table*}
In contrast to splitting networks at the last FC layer, the multi-task deep CNN (MCNN) \cite{hand2017attributes} branches out to multiple groups at the mid-level convolutional layers for modeling the attribute correlations. Specifically, based on the assumption that many attributes are strongly correlated, MCNN divides all 40 attributes into 9 groups according to semantics, i.e., gender, nose, mouth, eyes, face, around head, facial hair, cheeks, and fat. For example, \emph{big nose} and \emph{pointy nose} are grouped into the `nose' category, and \emph{big lips}, \emph{lipstick}, \emph{mouth slightly open} and \emph{smiling} are clustered into the `mouth' category. Therefore, each group consists of similar attributes and learns high-level features independently. At the first two convolutional layers of MCNN, features are shared by all attributes. Then, MCNN branches out several forks corresponding to different attribute groups. That means each attribute group occupies a fork. At the end of the network, an FC layer is added to create a two-layer auxiliary network (AUX) to facilitate attribute relationships. AUX receives the scores from the trained MCNN and yields the final prediction results. Hence, MCNN-AUX models facial attribute relationships in three ways: (1) sharing the lowest layers for all attributes, (2) assigning the higher layers for spatially related attributes, and (3) discovering score-level relationships with the AUX network.

However, MCNN has a significant limitation: shared information at low-level layers may vanish after network splitting. One solution to overcome this limitation is jointly learning shared and attribute-specific features at the same level rather than in order of precedence.

Therefore, Cao et al. \cite{cao2018partially} design a partially shared structure based on MCNN, i.e., PS-MCNN. It divides all 40 attributes into 4 groups according to attribute positions, i.e., upper group, middle group, lower group, and whole image group. Note that the entire partition process is performed by hand, and this artificial grouping strategy can be regarded as the prior information based on human knowledge. The partially shared structure connects four attribute-specific networks (TSNets) corresponding to four different groups of attributes and one shared network (SNet) sharing features among all the attributes. Specifically, each TSNet learns features for a specific group of attributes. Meantime, SNet shares informative features with each task. In terms of the connection mode between the SNet and the TSNets, each layer of SNet receives additional inputs from the previous layers of TSNet. Then, features from SNet are fed into the next layers of shared and attribute-specific networks. At a certain level of PS-MCNN, both task-specific features and shared features are captured in different branches. In addition, shared features at a specific layer are closely related to the features of all of its previous layers. This connection mechanism contributes to informatively shared feature representations.

Apart from attribute correlations, Han et al. \cite{han2017heterogeneous} introduce the concept of attribute heterogeneity. They note that individual attributes could be heterogeneous concerning data type and scale, as well as semantic meaning. In terms of data type and scale, attributes can be grouped into ordinal vs. nominal attributes. For instance, if attributes \emph{age} and \emph{hair length} are ordinal, then attributes \emph{gender} and \emph{race} are nominal. Note that the main difference between ordinal and nominal attributes is ordinal attributes have an explicit ordering of their variables, whereas nominal attributes generally have two or more classes and there is no intrinsic ordering among the categories. In terms of semantic meaning, attributes such as \emph{age}, \emph{gender}, and \emph{race} are used to describe the characteristics of the whole face, and \emph{pointy nose} and \emph{big lips} are mainly used to describe the local characteristics of facial components. Therefore, these two categories of attributes are heterogeneous and can be grouped into holistic vs. local attributes for the prediction of different parts of a face image. Therefore, taking both the attribute correlation and heterogeneity into consideration, Han et al. design a deep multi-task learning (DMTL) CNN to learn shared features of all attributes and category-specific features of heterogeneous attributes. The shared feature learning naturally exploits the relationship among attributes to yield discriminative feature representations, whereas the category-specific feature learning aims to fine-tune the shared features towards the optimal estimation of each heterogeneous attribute category.

Note that existing multi-task learning methods make no distinction between low-level and mid-level features for different attributes. This is unreasonable because features at different levels of the network may have different relationships. Besides, the above methods share features across tasks and split layers that encode attribute-specific features by hand-designed network architectures. Such a manual exploration in the space of possible multi-task deep architectures is tedious and error-prone because possible spaces might be combinatorially large.

In light of this issue, Lu et al. \cite{lu2017fully} present the automatic design of compact multi-task deep learning architectures, with no need to artificially discover possible multi-task architectures. The proposed network learns shared features in a fully adaptive way, where the core idea is incrementally widening the current design in a layer-wise manner. During the training process, the adaptive network starts with a thin multi-layer network (VGG16) and dynamically widens via a top-down layer-wise model widening strategy \cite{tropp2006SMOP}. It decides with whom each task shares features in each layer, yielding corresponding branches in this layer. Finally, the number of branches at the last layer of the model is equal to that of the attribute categories to be predicted. Consequently, this training scheme considers both task correlations and the complexity of the model for facilitating task grouping decisions at each layer of the network. Therefore, the fully-adaptive network allows us to estimate multiple facial attributes in a dynamic branching procedure through its self-constructed architecture and feature sharing strategy.

To summarize, holistic methods take the entire face images as inputs and mainly work on exploring attribute relationships. Many methods design various network architectures to model the correlations among different attributes. The key to this idea is learning shared features at low-level layers and attribute-specific features at high-level layers. Thus, holistic FAE methods need to address two main problems: one is assigning different layers for learning corresponding features with different characteristics, and another is learning more discriminative features though discovering attribute relationships under customized networks. What can be observed from contemporary research is that attribute grouping by hand has become a prevalent scheme in holistic FAE. We expect that an automatic attribute grouping strategy would attract more attention in future work, and it should adaptively learn proper group partition criteria and adjust them according to models' performance during the training.

\section{State-of-the-art Facial Attribute Manipulation Methods}\label{section4}
In this section, we provide an overview of model-based FAM methods and extra condition-based FAM methods in terms of algorithms, network architectures, advantages and disadvantages. The summary of this overview is provided in Table~\ref{Deep FAM}.

\subsection{Model-based Deep FAM Methods}
Model-based methods map an image in the source domain to the target domain and then distinguish the generated target distribution with the real target distribution under the constraint of an adversarial loss. Therefore, model-based methods are greatly task-specific and have excellent performance in yielding photorealistic facial attribute images.

Li et al. \cite{li2016IDaware} first propose a DIAT model following the standard paradigm of model-based methods. DIAT takes unedited images as inputs to generate target facial images with an adversarial loss and  an identity loss. The first loss ensures to obtain desired attributes, and the second encourages the generated images to have the same or similar identity as the input images. Zhu et al. \cite{zhu2017unpaired} add an inverse mapping from the target domain to the source domain based on DIAT and propose a CycleGAN, where the two mappings are coupled with a cycle consistency loss. This design is based on the intuition that if we translate from one domain to the other and back again, we should arrive where we start. Based on CycleGAN, Liu et al. \cite{liu2017unsupervised} propose a UNIT model that maps the pair of corresponding images in the source and the target domains to the same latent representation in a shared latent space. Each branch from one of the domains to the latent space performs an analogous CycleGAN operation.

However, all of the above methods directly operate on the entire face image. That means when a certain attribute is edited, the other relevant attributes may also be changed uncontrollably.

Therefore, to modify attribute-specific face areas and keep the other parts unchanged, Shen et al. \cite{shen2017learning} present learning residual images, which are defined as the difference between images before and after attribute manipulation. In this way, face attributes can be efficiently manipulated with modest pixel modification over the attribute-specific regions. They design a ResGAN consisting of two image transformation networks and a discriminative network to learn residual representations of desired attributes. Specifically, two image transformation networks, denoted as $G_0$ and $G_1$, first take two images with opposite attributes as inputs in turn and then perform the inverse attribute manipulation operation for outputting residual images. Subsequently, the obtained residual images are added to the original input images, yielding the final outputs with manipulated attributes. In the end, all these images, i.e., the two original input images and the two images from the transformation networks, are fed into the discriminative network, which classifies these images into three categories: images generated from the two transformation networks, original images with positive attribute labels, and original images with negative attribute labels. Note that $G_0$ and $G_1$ constitute a dual learning cycle. Given an image with a negative attribute label, $G_0$ synthesizes the desired attribute, and $G_1$ removes the corresponding attribute that is generated by $G_1$. Then, $G_1$'s output is expected to have the same attribute label as the original given image. The experiments demonstrate that such a dual learning process is beneficial for the generation of high-quality images, and residual images could enforce the attribute manipulation process to focus on the local areas where attributes show up. Therefore, ResGAN is able to generate attractive images especially on local facial attributes.

However, model-based methods can only edit an attribute during a training process with a set of corresponding model parameters. The whole manipulation is only supervised by discriminating real or generated images with the adversarial loss. That means when multiple attributes need to be changed, multiple training processes are inevitable, resulting in significant time consumption and computation costs.

In contrast, manipulating facial attributes with extra conditions is a more prevalent approach since multiple attributes can be edited through a single training process. Hence, extra condition-based methods attract more attention from researchers, where extra attribute vectors and reference exemplars are taken as input conditions. Specifically, attribute vectors can be concatenated with the latent image codes to control facial attributes, whereas reference exemplars exchange specific attributes with the to-be-manipulated images in the image-to-image translation framework. More details about the extra condition-based deep FAM methods are introduced below.
\begin{table*}[!t]
\renewcommand\arraystretch{1.5}
\begin{center}
\caption{State-of-the-art Facial Attribute Manipulation Approaches.}\label{Deep FAM}
\resizebox{\textwidth}{!}{
\begin{tabular}{|c|c|c|c|}
  \hline
  Approaches & Algorithms & \tabincell{c}{Network \\Architectures} & Datasets\\
  \hline\hline
  \multicolumn{4}{|c|}{\emph{MODEL-BASED METHODS}} \\\hline
  DIAT \cite{li2016IDaware} (arxiv1610)& \tabincell{c}{Transferring input images to each reference attribute label while keeping the same \\ or similar identity for Identity-Aware Transfer (DIAT) of facial attributes} & GAN & CelebA \\
  \hline
  InfoGAN \cite{chen2016infogan} (NIPS2016)& \tabincell{c}{Maximizing mutual information for interpretable representations \\and discovering visual concepts of facial attributes} & GAN & CelebA\\
  \hline
  UNIT \cite{liu2017unsupervised} (NIPS2017)& \tabincell{c}{Proposing an UNsupervised Image-to-Image Translation (UNIT) \\ framework under a shared-latent assumption } & GAN$+$VAE & CelebA \\
  \hline
  Residual Image \cite{shen2017learning} (CVPR2017)& \tabincell{c}{Learning residual images to avoid \\entire face operation with redundant irrelevant information } & GAN & CelebA\\
  \hline
  Wang et al. \cite{wang2018weakly} (WACV2018)& \tabincell{c}{Combining a perceptual content loss and two adversarial losses to guarantee \\ the global consistency for producing more realistic images} & GAN & \tabincell{c}{CelebA \\ LFW }\\
  \hline
  SG-GAN \cite{zhang2018sparsely} (ACMMM2018)& \tabincell{c}{Constructing a sparsely grouped generative adversarial networks (SG-GAN) \\ in the sparsely grouped datasets where most training data is mixed and a few are labelled} & GAN & \tabincell{c}{CelebA}\\
  \hline\hline
  \multicolumn{4}{|c|}{\emph{EXTRA CONDITION-BASED METHODS}} \\\hline
  \multicolumn{4}{|l|}{\emph{Conditioned on attribute vectors}} \\\hline
  VAE/GAN \cite{larsen2016autoencoding} (ICML2016)& \tabincell{c}{Using learned feature representations in the GAN discriminator as basis\\ for the VAE reconstruction objective} & GAN$+$VAE & LFW\\
  \hline
  CVAE \cite{yan2016attribute2image} (ECCV2016)& \tabincell{c}{Learning a layered foreground-background generative \\ conditional variational auto-encoder for complex images} & VAE & LFW \\
  \hline
  IcGAN \cite{perarnau2016invertible} (NIPSW2016)& Combining an encoder with a cGAN for obtaining Invertible cGAN (IcGAN) & GAN$+$VAE & CelebA \\
  \hline
  Fader Network \cite{lample2017fader} (NIPS2017)& \tabincell{c}{Disentangling the salient information of face images and the values of attributes \\ directly in the latent space for modifying facial attributes continuously} & AE & CelebA \\
  \hline
  cCycleGAN \cite{lu2018attribute} (ECCV2018)& \tabincell{c}{Extending the cycleGAN \cite{zhu2017unpaired}conditioned on facial \\ attributes with the cycle consistency loss} & GAN & CelebA \\
  \hline
  StarGAN \cite{choi2017stargan} (CVPR2018)& \tabincell{c}{Constructing a StarGAN for multiple domain image-to-image translations} & GAN & CelebA \\
  \hline
  CRGAN \cite{li2018facial} (Springer JCST2018)& \tabincell{c}{Introducing recycle reconstruction loss to maintain personal facial identity \\ and directly learning facial transformation with attribute annotations} & GAN & \tabincell{c}{CelebA}\\
  \hline
  SaGAN \cite{zhang2018generative} (ECCV2018)& \tabincell{c}{Introducing a spatial attention mechanism for \\only modifying the attribute-specific region and keep the remains unchanged} & GAN & \tabincell{c}{CelebA\\LFW} \\
  \hline
  \multicolumn{4}{|l|}{\emph{Conditioned on reference exemplars}} \\\hline
  Gene-GAN \cite{zhou2017genegan} (BMVC2017)& \tabincell{c}{Recombing the latent representation information of two \\ paired attribute images for swapping specific attributes} & GAN &  \tabincell{c}{CelebA}\\
  \hline
  DNA-GAN \cite{xiao2017dna} (ICLRW2018)& \tabincell{c}{Learning disentangled representations from multi-attribute images\\ by annihilating and swapping operations to achieve the attribute manipulation} & GAN & CelebA \\
  \hline
  ELEGANT \cite{xiao2018elegant} (ECCV2018)& \tabincell{c}{Exchanging Latent Encoding with GAN for Transferring Multiple Face Attributes (ELEGANT) \\ and doing image generation by exemplars as well as producing high-quality generated images} & GAN$+$VAE& CelebA \\
  \hline
  EGSC-IT \cite{ma2018exemplar} (ICLR2019) & \tabincell{c}{Constructing an exemplar guided semantically consistent image-to-image translation (EGSC-IT) network \\to control the translation process under exemplar images in the target domain.} & GAN$+$VAE & CelebA\\
  \hline
\end{tabular}}
\end{center}
\end{table*}
\subsection{Extra Condition-based Deep FAM Methods}
Deep FAM methods conditioned on extra attribute vectors alter desired attributes with given conditional attribute vectors, such as one-hot vectors indicating the presence of corresponding facial attributes. During the training process, the conditional vectors are concatenated with the to-be-manipulated images in latent encoding spaces. Moreover, conditional generative frameworks dominate the model construction of deep FAM. Various efforts have been made to edit facial attributes based on autoencoders (AEs), VAEs, and GANs.

Zhang et al. \cite{zhang2017age} propose a conditional adversarial autoencoder (CAAE) for age progression and regression. CAAE first maps a face image to a latent vector through an encoder. Then, the obtained latent vector concatenated with an age label vector is fed into a generator for learning a face manifold. The age label condition controls altering the age. Meanwhile, the latent vector ensures that the personalized face features are preserved. Yan et al. \cite{yan2016attribute2image} introduce a conditional variational autoencoder (CVAE) to generate images from visual attributes. CVAE disentangles an image into the foreground and the background parts, where each part is combined with the defined attribute vector. Consequently, the quality of generated complex images can be significantly improved when the foreground areas attract more attention. Perarnau et al. \cite{perarnau2016invertible} propose an invertible conditional GAN (IcGAN) to edit multiple facial attributes with determined specific representations of generated images. Given an input image, IcGAN first learns a representation consisting of a latent variable and a conditional vector via an encoder. Then, IcGAN modifies the latent variable and conditional vector to regenerate the original input image through the conditional GAN \cite{mirza2014conditional}. In this way, by changing the encoded conditional vector, IcGAN can achieve arbitrary attribute manipulation.

Apart from autoencoders, VAEs, GANs, and their variants, Larsen et al. \cite{larsen2016autoencoding} combine the VAE and the GAN into a unified generative model, VAE/GAN. In this model, the GAN discriminator learns feature representations taken as the basis of the VAE reconstruction objective, which means that the VAE decoder and the GAN generator are collapsed into one by sharing parameters and joint training. Hence, this model consists of three parts: the encoder, the decoder, and the discriminator. By concatenating attribute vectors with features from these three components, VAE/GAN performs better than either plain VAEs or GANs.

Recently, taking the multiple attribute manipulation as a domain transfer task, Choi et al. \cite{choi2017stargan} propose a StarGAN to learn mappings among multiple domains with only a single generator and a discriminator trained from all domains. Each domain corresponds to an attribute and the domain information can be denoted by one-hot vectors. Specifically, the discriminator first distinguishes the real and the fake images and classifies the real images to their corresponding domains. Then, the generator is trained to translate an input image into an output image conditioned on a target domain label vector, which is generated randomly. As a result, the generator is capable of translating the input image flexibly. In summary, StarGAN takes the domain labels as extra supervision conditions. This operation makes it possible to incorporate multiple datasets containing different types of labels simultaneously.

However, all the above methods edit multiple facial attributes simultaneously by discretely changing multiple values of attribute vectors. None of them can alter facial attributes continuously.

In light of this,  Lample et al. \cite{lample2017fader} present a Fader network using continuous attribute values to modify attributes through sliding knobs, like faders on a mixing console. For example, one can gradually change the values of \emph{gender} to control the transition process from \emph{man} to \emph{woman}. Fader network is composed of three components: an encoder, a decoder, and a discriminator. With an image-attribute pair as the input, Fader network first maps the image to the latent representation by its encoder and predicts the attribute vector by its discriminator. Then, the decoder reconstructs the image through the learned latent representation and the attribute vector. During testing, the discriminator is discarded, and different images with various attributes can be generated with different attribute values.

Note that all the above methods edit attributes over the whole face images. Hence, attribute-irrelevant details might also be changed. To address this issue, Zhang et al. \cite{zhang2018generative} introduce the spatial attention mechanism into GANs to locate attribute-relevant areas and propose a SaGAN for manipulating facial attributes more precisely. SaGAN follows the standard adversarial learning paradigm, where a generator and a discriminator play a min-max game. To keep attribute-irrelevant regions unchanged, SaGAN's generator consists of an attribute manipulation network (AMN) and a spatial attention network (SAN). Given a face image, SAN learns a spatial attention mask where attribute-relevant regions have non-zero attention values. In this way, the region where the desired attribute appears can be located. Then, AMN takes the face image and the attribute vector as inputs, yielding an image with the desired attribute in the specific region located by SAN.

Rather than taking the attribute vectors as extra conditions, deep FAM methods conditioned on reference exemplars consider exchanging specific attributes with the to-be-manipulated images in the image-to-image translation framework. Note that these reference images do not need to have the same identity as the original to-be-manipulate images, and all the generated attributes are present in the real world. In this way, more specific details that appear in the reference images can be explored to generate more realistic images.

Zhou et al. \cite{zhou2017genegan} first design a GeneGAN to achieve the basic reference exemplar-based facial attribute manipulation. Given an image, it is encoded into two complement codes: attribute-specific codes and attribute-irrelevant codes. By exchanging the attribute-specific codes and preserving the attribute-irrelevant codes, desired attributes can be transferred from the reference exemplar image to the to-be-manipulated image.

Considering that GeneGAN only transfers one attribute in a single manipulation process, Xiao et al. \cite{xiao2018elegant} construct an ELEGANT model to exchange latent encodings for transferring multiple facial attributes by exemplars. Specifically, since all the attributes are encoded in the latent space in a disentangled manner, one can exchange the specific part of encodings and manipulate several attributes simultaneously. Besides, the residual image learning and the multi-scale discriminators for adversarial training enable the proposed model to generate high-quality images with more delicate details and fewer artifacts. At the beginning of training, ELEGANT receives two sets of training images as inputs, i.e., a positive set and a negative set, which do not need to be paired. Second, an encoder is utilized to obtain the latent encodings of both positive and negative images. Then, if the $i$-th attribute is required to be transferred, the only step is to exchange the $i$-th element in the latent encodings of positive and negative images. Once the encoding step is finished, ELEGANT constructs an image generator that consists of a decoder and the encoder from the previous step to decode recombined latent encodings into images. Finally, two discriminators with identical network structures work at different scales to obtain manipulated attribute images.

\section{Additional Related Issues}\label{section6}
\subsection{Imbalance Learning in Facial Attribute Analysis}\label{section6_1}
Face attribute data exhibits an imbalanced distribution in terms of different categories. It is normally called the class-imbalance issue, which means in a dataset, some of the facial attribute classes have a much higher number of samples than others, corresponding to the majority class and minority class \cite{haixiang2017learning}, respectively. For example, the largest imbalance ratio between the minority and majority attributes in CelebA dataset is 1:43. Learning from such imbalanced facial attribute labels can lead to biased classifiers, which tend to favor the majority and fail to discriminate the features learned from the minority. Even in the extreme case, the learned classifiers can hardly identify the minority samples.

One typical scheme to solve this problem is using an assumed balanced target distribution to guide the imbalanced source distribution by weighting objective functions. MOON \cite{rudd2016moon} weights the back-propagation error in a cost-sensitive way. A probability is assigned to each class by counting the relative numbers of positive and negative samples for both source and target domains. Then, these probabilities could be used as weights to incorporate the distribution discrepancy into the loss function.

However, MOON overlooks the label imbalance over each batch, which means that the batch-wise training scheme of deep networks is not fully utilized. In light of this,  AttCNN \cite{hand2018doing} proposes a selective learning algorithm to address the distribution discrepancy at the batch level. If the original batch in the source domain has more positive samples and fewer negative samples than the target distribution, the selective learning algorithm resamples a random subset from the positive instances. Meanwhile, it proportionally weights the negative counterparts to match the target distribution. By aligning the distributions between the source and target domains in each batch, AttCNN yields the state-of-the-art class-imbalance attribute prediction performance.

In addition, another more frequently used scheme for class-imbalance learning is data resampling for deep FAE methods. Huang et al. \cite{huang2016learning} adopt the resampling strategy, namely large margin local embedding (LMLE), and formulate a quintuple sampling term associated with the triple-header loss. LMLE enforces the preservation of locality across clusters and the discrimination between classes. Then, a fast cluster-wise kNN algorithm is executed, followed by a local large margin decision. In this way, LMLE learns embedded features that are discriminative enough without any possible local class imbalance. On this basis, Huang et al. further propose a rectified version of LMLE \cite{huang2019deep}, i.e., cluster-based large margin local embedding (CLMLE). CLMLE designs a loss to preserve the inter-cluster margins both within and between classes. In contrast to LMLE enforcing the Euclidean distance on a hypersphere manifold, CLMLE adopts angular margins enforced between the involved cluster distributions and uses spherical k-means for obtaining \emph{K} clusters with the same size, which contributes to better performance.

On the other hand, Dong et al. \cite{dong2017CRL} take an online regularization strategy to address the facial attribute based class-imbalance issue. In detail, they exploit a batch-wise incremental hard mining on minority attribute classes, and formulate a class rectification loss (CRL) based on the mined minority examples. For the hard mining strategy, researchers first provide the profiles of hard positives and hard negatives for the minority. Then, according to the predefined profiles and model, they select \emph{K} hard positives (or hard negatives) as the bottom-\emph{K} (or top-\emph{K}) scores on the minority class for a specific attribute. This process is executed at the batch level and incrementally over subsequent batches. Such batch-wise incremental hard mining guarantees CRL strong class-imbalance learning ability and satisfactory attribute estimation performance.

\subsection{Relative Attribute Ranking in Facial Attribute Analysis}

Relative attribute learning aims to formulate functions to rank the relative strength of attributes \cite{chen2014predicting}, which can be widely applied in object detection \cite{fan2013relative}, fine-grained visual comparison \cite{shi2018fine}, and facial attribute estimation \cite{li2018landmark}. The general insight in this line of work is learning global image representations in a unified framework \cite{lampert2009learning,parikh2011relative} or capturing part-based representations via pretrained part detectors \cite{bourdev2011describing,sandeep2014relative,zhang2014panda} . However, the former ignores the localizations of attributes, and the latter ignores the correlations among attributes. Consequently, both the two might collapse the performance of relative attribute ranking.

Xiao et al. \cite{xiao2015discovering} first propose automatically discovering the spatial extent of relevant attributes by establishing a set of visual chains indicating the local and transitive connections. In this way, the locations of attributes can be learned automatically in an end-to-end way. Although no pretrained detectors are used, the optimization pipeline still contains several independent modules, resulting in a suboptimal solution.

To tackle this issue, Singh et al. \cite{singh2016end} construct an end-to-end deep CNN for simultaneously learning features, localizations, and ranks of facial attributes with weakly supervised pair-wise images. Specifically, given pairs of training images ordered according to the relative strength of an attribute, two Siamese networks receive these images, where each takes one of a pair as input and builds a single branch. Each branch contains two components: the spatial transformer network (STN), which generates image transformation parameters for localizing the most relevant regions, and the ranker network (RN), which outputs the predicted attribute scores. The qualitative experiment results over LFW-10 dataset show excellent performance in attribute region localization and ranking accuracy.

To model the pair-wise relationships between images for multiple attributes, Meng et al. \cite{meng2018efficient} construct a graph model, where each node represents an image and edges indicate the relationships between images and attributes, as well as between images and images. The overall framework consists of two components: the CNN for extracting primary features of the node images, and the graph neural network (GNN) for learning the features of edges and following updates. Thus, the relationships among all the images are modeled by an fully-connected graph over the learned CNN features. Then, a gated recurrent unit (GRU) takes the node and its corresponding information as inputs and outputs the updated node. As a result, the correlations among attributes can be learned by using information from the neighbors of the node, as well as by updating its state based on the previous state.

\subsection{Adversarial Robustness in Facial Attribute Analysis}
Adversarial images, which are generated from the network topology, training process, and hyperparameter variation by adding slight artificial perturbations, can be used as inputs of deep facial attribute analysis models. By classifying the original inputs correctly and  misclassifying the adversarial inputs, the robustness of models can be improved. Szegedy et al. \cite{szegedy2013intriguing} first propose that neural networks can be induced to misclassify an image by carefully chosen perturbations that are imperceptible to human. Following this work, the study of adversarial images is entering the horizons of researchers.

Rozsa et al. \cite{rozsa2017facial} induce small artificial perturbations on existing misclassified inputs to correct the results of attribute classification. Specifically, the adversarial images are generated over a random subset of CelebA dataset via the fast flipping attribute (FFA) technique. FFA algorithm leverages the back-propagation of the Euclidean loss to generate adversarial images. During this process, it flips the binary decision of the deep network without ground-truth labels. Through the robustness analysis, FFA has better performance in generating more adversarial examples than the existing fast gradient sign (FGS) method \cite{goodfellow2015explaining} on the designed separate attribute networks \cite{rozsa2016facial}. Moreover, FFA algorithm is extended to an iterative version, namely iterative FFA, to ensure the use for multi-objective networks, e.g., MOON \cite{rudd2016moon}. The experiments demonstrate that the quality of adversarial examples of iterative FFA is more satisfactory than its base version, and iterative FFA can flip attribute prediction results more frequently. Despite the promising performance of these two types of FFAs, several attributes still could not be flipped over on separately trained deep models.

In addition, attribute anonymity, which conceals specific facial attributes that an individual does not want to share, is another adversarial robustness related task. When hiding corresponding attributes, the remaining attributes should be maintained, and the visual quality of images should not be damaged. Chhabra et al. \cite{chhabra2018anonymizing} achieve this basic target by adding adversarial perturbations to an attribute preservation set and an attribute suppression set. Consequently, the prediction of a specific attribute from the true category can be classified into a different target category.

In summary, the study of adversarial robustness contributes to improving the representational stability of current deep FAE algorithms. Additionally, due to the attack of adversarial examples, the robustness of deep facial attribute analysis models is moving towards a promising direction.
\section{Challenges and Opportunities}\label{section7}
Despite the promising performance of many algorithms in deep facial attribute analysis, there are still several challenging issues that deserve more attention. On the other hand, these challenges also bring hopeful opportunities for the development of this field. Therefore, in this section, we discuss challenges and future opportunities for both deep FAE and FAM, from the perspectives of databases, algorithms, and real-world applications.
\subsection{Discussion of Facial Attribute Estimation}
\subsubsection{Data}
The development of deep neural networks makes FAE a data-driven task. That means large numbers of samples are required for training deep models to capture attribute-relevant facial details. However, contemporary studies suffer from insufficient training data. In this case, deep neural networks would easily fit the data characteristics contained only in a small number of images and have degraded performance. In the following, taking two commonly used datasets as examples (i.e., CelebA and LFWA), we analyze the data challenges that exist in current facial attribute databases from the perspectives of data sources, data quality, and imbalanced data, respectively.

First, from the perspective of data sources, CelebA collects face data and attribute labels from the celebrities, and the samples of LFWA come from online news. There is no doubt that these databases are inherently biased and do not match the general data distributions in the real world. For example, the \emph{bald} attribute corresponds to a small number of samples in CelebA, but in the real world, it is a common attribute among ordinary people. Hence, more complementary facial attribute datasets that cover more real-world scenarios and a wider range of facial attributes need to be constructed in the future. An earlier work \cite{wang2016walk} has made an attempt to extract images from the real-world outdoor videos, i.e., Ego-Humans dataset. However, it contains more pedestrian attributes, and only several facial attributes are predicted. Nevertheless, we believe that this dataset provides an inspired idea for collecting more facial attribute-relevant images from videos in real-world scenes \cite{wiles2018self}.

Furthermore, Hand et al. \cite{hand2018predicting} have made the first attempt to estimate facial attributes in videos. They use weakly labeled data in YouTube Faces Dataset (with attribute labels) to keep attribute prediction consistent and accurate in videos, by imposing a temporal coherence constraint and a motion-attention mechanism. The temporal coherence constraint ensures the response invariability between video frames by transferring responses from labeled frames to unlabeled ones. Meanwhile, the motion-attention mechanism enforces their model to focus on face parts through exploring the motion relationship between labeled and unlabeled frames. On the one hand, this research significantly highlights the importance of temporal and motion factors when designing video-based deep FAE models. On the other hand, it also expresses the expectation for labeling new video datasets with facial attributes in future study.

Second, from the perspective of data quality, most faces in CelebA and LFWA are frontal and aligned images with high quality \cite{hand2018doing}. However, real-world data always have low-quality, partially visible images with various illumination and poses. Thus, attribute prediction models trained on these images could hardly learn representative features of real-world data. Therefore, we expect that more adequate real-world training data would come out to strengthen the estimation abilities of future attribute classifiers.

Finally, for CelebA, LFWA, or real-world face images, imbalanced data would induce attribute estimation models to pay more attention to learning the features of majority samples. Consequently, learned biased attribute classifiers could not identify the minorities in some extreme cases. Although many efforts have been made to solve this class-imbalance learning issue from the perspective of algorithms, as mentioned in Section~\ref{section6_1}, data support is still an urgent need.

Besides, the test datasets (i.e., target domains), may have different distributions from the training datasets (i.e., source domains). It is generally called domain adaption issue, which can be taken as a distribution imbalance. That means once the source data have a particular property, the given target domain would not always follow the same pattern. Therefore, such a discrepancy between data distributions would negatively impact the generalization ability over unseen test data and lead to significant performance deterioration.

Therefore, on the one hand, we anticipate that more available facial attribute images can be released to capture discriminative features of majority and minority samples equally well in terms of class-imbalance data. On the other hand, more algorithms are expected to be developed to solve the domain adaption issue in attribute estimation.

\subsubsection{Algorithms}
As mentioned before, part-based deep FAE methods and holistic deep FAE methods develop in parallel. The former pays more attention to locating attributes, and the latter concentrates more on modeling attribute relationships. Below, we provide the main challenges from the perspective of algorithms and analyze the future trends for both types of methods.

For the part-based methods, earlier methods draw support from existing part detectors to discover facial components. However, these detected parts of faces are coarse and attribute-independent. They only distinguish the whole face from the other face-irrelevant parts, such as the background in an image. Considering that existing detectors are not customized for deep FAE, some researchers begin to seek help from other face-related auxiliary tasks, which focus more on facial details rather than the whole face. There are also some studies that utilize labeled key points to partition facial regions. However, well-labeled facial images are not always available in real-world applications, and the performance of auxiliary tasks would limit the accuracy of the downstream classification task.

We believe that an end-to-end strategy would dominate future part-based deep FAE algorithms, where the attribute-relevant regions and the corresponding prediction can be yielded in a unified framework \cite{fukui2019attention}. Ding et al. \cite{ding2017deepCascadeFAE} have attempted to tackle this issue, but learning a region for each attribute is cumbrous and computationally expensive. This is because several attributes might appear in the same region of a face. 

In addition, part-based methods show great superiority when dealing with data under in-the-wild environmental conditions, such as illumination variations, occlusions, and non-frontal faces. Through learning the locations of different attributes, part-based methods integrate the information from non-occluded areas to predict attributes in occluded areas. Mahbub et al. \cite{mahbub2018segment} address this issue by partitioning facial parts manually according to key points. However, such annotations are not always available. Attempting to integrate these non-occluded areas adaptively is becoming a future trend. Besides, Mahbub et al. \cite{mahbub2018segment} test their model's attribute estimation performance on partial faces by adding occlusions artificially over original databases, but this operation is not normative for the test protocol. Therefore, the lack of data under the in-the-wild conditions is still a challenge for training deep FAE networks in the wild environment.

For holistic methods, state-of-the-art approaches design networks with different architectures for sharing common features and learning attribute-specific features at different layers. However, these methods define attribute relationships to design networks by grouping attributes manually, which can be taken as extra prior information. Since different individuals might give different attribute partitions according to locations or semantics, it is difficult to determine that which facial attribute groups are suitable and optimal. Therefore, how to discover attribute relationships adaptively in the training process, without given prior information artificially, should be the focus of future works.

In addition, facial attributes have been taken as auxiliary and complementary information for many face-related tasks, such as face recognition \cite{kumar2009attribute,rudd2016moon,taherkhani2018deep}, face detection \cite{ranjan2017hyperface}, and facial landmark localization \cite{zhuang2018DeepTranfer}. Kumar et al. \cite{kumar2009attribute} first introduce the concept of `attribute' to facilitate face verification by compact visual descriptions and low-level attribute features. In contrast, Rudd et al. \cite{rudd2016moon} utilize the mixed objective optimization network with the Euclidean loss to learn deep attribute features for promoting facial verification. Experiments illustrate that despite only 40 attributes being used, the work of Rudd et al. \cite{rudd2016moon} still performs better than that of Kumar et al. \cite{kumar2009attribute}, which extracts features of 73 facial attributes.

Apart from employing features learned by attribute prediction to assist face recognition, joint and incorporative learning of facial attribute relevant tasks can further enhance their respective robustness and performance by discovering complementary information. For example, considering the inherent dependencies of face-related tasks, Zhuang et al. \cite{zhuang2018multi} design a cascaded CNN for simultaneously learning face detection, facial landmark localization, and facial attribute estimation under a multi-task framework to improve the performance of each task. They further attempt to perform joint face recognition and facial attribute estimation when taking the relationship between identities and attributes into account. Therefore, it is reasonable to believe that the combination of different face-related tasks is becoming a promising research direction due to the complementary relationships among them.
\subsubsection{Applications}
Various viewpoints of the same person are difficult challenges for maintaining the identity-attribute consistency in deep FAE methods. On the one hand, such viewpoint diversification helps to learn richer features from the same person. On the other hand, images of different viewpoints might differ in attributes even from the same identity. For example, the side face images might yield different prediction results with the front face images for the \emph{high cheekbones}, as the side face images do not emphasize this attribute.

Therefore, attribute inconsistency becomes a severe problem in various viewpoints for the same identity. Lu et al. \cite{lu2018attribute} propose a probabilistic confidence criterion to address this inconsistency issue. Specifically, this criterion first extracts the most confident face image for each subject, and then it chooses the result corresponding to the highest confidence as the final prediction of each attribute concerning each subject. However, filtering the most confident image via relevant criteria might not be the most optimal strategy, because the features from all images with different views are not taken full advantage of in making the favorable estimation.

Nowadays, digital mobile devices contain considerable amounts of valuable personal information, such as bank accounts and private emails \cite{samangouei2017facial}. These personal details make these devices the targets of various attacks. Hence, biological characteristics, such as fingerprints and irises \cite{trokielewicz2019iris}, have been widely used as device passwords for further protecting the privacy information of users. This technique is called biometric verification. Recently, an increasing number of biometric verification based algorithms have emerged as a solution for continuous authentication on mobile devices. Many researchers have committed to designing active authentication algorithms based on face biometrics. For example, studies in \cite{fathy2015face,gunther20132013,hadid2007face} detect faces through camera sensor images and further extract low-level features for the authentication of smartphone users.

Considering that facial attributes contain more detailed characteristics than the full face, we believe that facial attributes would bring new opportunities for biometric identification in real-world applications. Samangouei et al. \cite{samangouei2017facial} have attempted the active authentication of mobile devices by facial attributes. A set of binary attribute classifiers are trained to estimate whether attributes are present in images of the current user in a mobile device. Consequently, the authentication can be implemented by comparing the recognized attributes with the originally enrolled attributes.

However, Samangouei et al. \cite{samangouei2017facial} extract traditional features, such as the LBP feature, which are not task-specific for attribute estimation and less discriminative than deep features. To some extent, these traditional features and SVM classifiers balances the verification accuracy and mobile performance, whereas other methods with satisfactory performance might have tremendous computation or memory costs.

Therefore, future challenges mainly lie in two aspects. The first is to better apply facial attributes for mobile device authentication. The second is exploring more discriminative deep features and classifiers under the constraints of the trade-off between verification accuracy and mobile performance. Nevertheless, we expect that facial attributes would contribute to further advance the progress of biometric verification on digital mobile devices.

\subsection{Discussion of Facial Attribute Manipulation}
\subsubsection{Data}
In this section, we start with the problems of current FAM databases and analyze the challenges and the opportunities related to data sources. Then, we express an expectation for the video data type, as we have done in the discussion of facial attribute prediction. Finally, taking the performance metrics into account, we believe that future deep FAM methods need to establish a unified standard for evaluating their experiment results.

First, in terms of data sources, note that almost all deep FAM algorithms are trained over CelebA database, while very few of them also use LFW dataset. The data sources are extremely inadequate, and facial attributes that can be manipulated are considerably limited. For 40 annotated attributes, only several notable attributes (e.g., \emph{hair colors} \cite{li2018facial}, \emph{glasses} \cite{chen2016infogan}, and \emph{smiling} \cite{xiao2018elegant}) can achieve satisfactory performance. Such limitation could cause a degradation in performance when manipulating various attribute types. Therefore, we expect that more high-quality facial attribute databases could be released and that more kinds of facial attributes could be manipulated in the future.

Second, from the perspective of the data type, FAM on the video data still has not been studied. Manipulating video facial attributes requires models to yield lifelike details.
When faces change with the frames of videos, models can still locate the to-be-manipulated areas precisely and keep the consistency of attribute manipulation for the same identity. Nevertheless, this task is valuable in many entertainment situations in the real world, such as beauty makeup videos. The hair colors in the videos might be varied according to users' preference. However, to date, there is no available large-scale video data for training video-based attribute manipulation models. The possible reasons might be that it is difficult to track and annotate facial attributes in large-scale videos due to spatial and temporal dynamics \cite{saito2017temporal}, and the quality of video data could have significant effects on such a synthesis task. We expect that the focus will be shifted to collect and annotate video data with facial attributes for promoting the video-based deep FAM task further.

Finally, from the perspective of performance metrics, as mentioned in Section~\ref{section5}, contemporary research either evaluates generated images by statistical surveys or seeks help from other face-related tasks, such as attribute estimation and landmark detection. Unified and standard metric systems have not yet formed in terms of qualitative and quantitative analyses. We expect that the metrics of deep FAM methods could be well developed and establish a relatively unified rule in the future.
\subsubsection{Algorithms}
State-of-the-art deep FAM methods can be grouped into two categories: model-based methods and extra condition-based methods. Model-based methods tackle an attribute domain transfer issue and use the adversarial loss to supervise the process of image generation. Extra condition-based methods alter desired attributes with given conditional attributes concatenated with to-be-manipulated images in encoding spaces. The main difference between the two types of methods is whether extra conditions are required.

Model-based methods take no extra conditions as inputs, and one trained model only changes one corresponding attribute. This strategy is task-specific and helps to generate more photorealistic images, but it is difficult to guarantee attribute-irrelevant details are unchanged due to its operation based on the whole image directly. Few methods focus on this issue, except for ResGAN proposed by Shen et al. \cite{shen2017learning}. However, ResGAN generates residual images for locating attribute-relevant regions under the sparsity constraint. Such a constraint relies heavily on control parameters but not attributes themselves. Hence, how to design networks to synthesize desired photorealistic attributes, as well as keep other attribute-irrelevant details unchanged, is a significant challenge in the future. In addition, as multi-domain transfer has become a hot research topic \cite{liu2018unified,zhang2018xogan}, we expect that these novel domain transfer algorithms would migrate to deep FAM methods for yielding more appealing performance.

Extra condition-based methods take attribute vectors or reference exemplars as conditions. These algorithms edit facial attributes by changing values of attribute vectors or latent codes of reference exemplars. One advantage of this strategy is multiple attributes can be manipulated simultaneously by altering multiple corresponding values of conditions. However, the concomitant disadvantage is also inevitable. That is, these methods cannot change attributes continuously since the values of attribute vectors are edited discretely. We believe that this shortcoming can be solved by interpolation schemes \cite{berthelot2018understanding} or semantic component decomposition \cite{chen2019semantic} in the future. In addition, as mentioned before, reference exemplar based algorithms are becoming a promising research direction. More specific details that appear in reference images can be explored to generate more photorealistic images compared with merely altering attribute vectors manually.
\subsubsection{Applications}
Face makeup \cite{li2018beautygan,chang2018pairedcyclegan,cao2019makeup} and face aging \cite{suo2010compositional,nhan2019automatic,liu2019attribute} are two hot topics in deep FAM related applications. They have played important roles in mobile device entertainment (e.g., beauty cameras) and identity-relevant face verification. Compared with general FAM, they focus more on more subtle face attribute details. For face makeup, it concentrates more on makeup related attributes, such as the types of eyeshadows and the colors of lipsticks. The focus of studies lies on facial makeup transfer and removal \cite{chang2018pairedcyclegan,cao2019makeup}, where makeup transfer aims to map one makeup style to another for generating different makeup styles \cite{li2018beautygan}, and makeup removal performs an opposite process which cleans off the existing makeup and provides support to makeup-invariant face verification \cite{cao2019makeup}. In terms of face aging, it renders face images with a wide range of ages and keeps identity information insusceptible. Hence, this task can not only be applied to digital entertainment but also provide support to social safety, such as fugitive researches and cross-age identity verification. The most crucial issue in face aging is that there are no sufficient paired images for the same person at different ages \cite{liu2019attribute}. Recently, the development of deep learning has lead face makeup and face aging to promising results, and they have become important research branches independent of general deep FAM methods. We expect the development of these two branches would bring out a hopeful prospect of future real-world applications.

Besides, resolution limitation is another tough challenge in real-world facial manipulation. Existing methods only work well with a limited range of resolutions and under lab conditions. This limitation encourages combining face super-resolution with deep FAM algorithms. For example, Lu et al. \cite{lu2018attribute} propose a conditional version of CycleGAN \cite{zhu2017unpaired} to generate face images under the guidance of attributes for face super-resolution. Specifically, conditional CycleGAN takes a pair of low/high-resolution faces and an attribute vector extracted from the high-resolution one as inputs. Conditioned on attributes of the original high-resolution image, this model learns to generate a high-resolution version of the original low-resolution image. Moreover, Dorta et al. \cite{dorta2018gan} apply smooth warp fields to GANs for manipulating face images with very high resolutions through a deep network at a lower resolution. All these schemes inspire researchers to integrate state-of-the-art face super-resolution methods into attribute manipulation for achieving a win-win situation.

\subsection{Relationships between FAE and FAM}
In this section, we introduce the relationships between deep FAE and FAM. We believe the discussion about how the two tasks assist each other would guide future research to improve both algorithms.

For deep FAE, deep FAM can be taken as a vital scheme of data augmentation, where generated facial attribute images can significantly increase the amount of data used for training deep neural networks. Sufficient training data can reduce the risk of overfitting and further improve the prediction accuracy. Future works should work harder on improving the quality of generated images and synthesizing as many facial attribute details as possible. In this way, generated images would better support the training of deep FAE models.

For deep FAM, the result of attribute estimation can be a significant quantitative performance evaluation criterion. The deep FAE network used for evaluation has to be well trained on real images in advance and has to provide an accuracy baseline for all real facial attributes. Then, it works on the generated facial attribute images and yields another prediction accuracy over manipulated attributes. As a result, the accuracy gap between real images and generated images can reflect the performance of deep FAM algorithms.

Despite the mutual assistance builds a bridge between deep FAE and deep FAM methods, there are still some issues that need to be addressed for the two tasks. First, generated facial attribute images may not contain too much delicate facial information. In other words, there is still a gap between real and augmented generated images, which might damage the performance of attribute estimation. Hence, how to close this gap can be an essential future research direction for data augmentation in deep facial attribute analysis. Second, the performance of attribute estimation directly affects the evaluation results of facial attribute manipulation. Therefore, how to balance the metric with the prediction performance is another challenge. We expect that deep FAE methods and deep FAM methods can strengthen their cooperation to significantly improve each others' performance in the future.

\section{Conclusion}\label{section8}
As one type of important semantic features describing the visual properties of face images, facial attributes have received considerable attention in the field of computer vision. The analyses targeting facial attributes, including facial attribute estimation (FAE) and facial attribute manipulation (FAM), have improved the performance of many real-world applications. This paper provides a comprehensive review of recent advances in both deep learning based FAE and FAM. The commonly used databases and metrics are summarized, and the taxonomies of state-of-the-art methods over both two issues have been created, together with their advantages and disadvantages. In addition, future challenges and opportunities are highlighted in terms of data, algorithms, and applications, respectively. We are looking forward to further studies that address these challenges and take these opportunities to promote the development of deep face attribute analysis.

\begin{acknowledgements} We thank the contributions of pioneer researchers in the field of deep learning based facial attribute analysis and other related fields. This work is supported in part by the State Key Development Program (Grant No. 2016YFB1001001), in part by the National Natural Science Foundation of China (NSFC) under Grant U1736119, and in part by the Fundamental Research Funds for the Central Universities under Grant DUT18JC06.
\end{acknowledgements}

%

\bibliographystyle{spmpsci}      
\bibliography{Survey}
%
%

\end{document}